\pdfoutput=1

\documentclass[11pt]{article}

\usepackage{subcaption}

\usepackage{caption}

\usepackage[final]{acl}

\usepackage{times}
\usepackage{latexsym}

\usepackage[T1]{fontenc}

\usepackage[utf8]{inputenc}

\usepackage{microtype}

\usepackage{inconsolata}
\usepackage{multirow}
\usepackage{placeins}

\usepackage{graphicx}
\usepackage{pifont}
\usepackage{float}
\usepackage{amsmath}
\usepackage{graphicx}
\usepackage[ruled,longend,linesnumbered]{algorithm2e}
\usepackage{hyperref}

\usepackage{todonotes}

\usepackage{booktabs}
\usepackage{colortbl}
\PassOptionsToPackage{table,xcdraw}{xcolor}
\usepackage{xcolor}

\newcommand{\cblock}[1]{\textcolor[rgb]{0, 0, 0}{{#1}}}
\newcommand{\cpass}[1]{\textcolor[rgb]{0, 0, 0}{{#1}}}
\newcommand{\canchor}[1]{\textcolor[rgb]{0, 0, 0}{{#1}}}
\newcommand{\name}{\textsc{APB}}

\title{\name: Accelerating Distributed Long-Context Inference by \\Passing Compressed Context Blocks across GPUs}

\author{
\\
 \textbf{Yuxiang Huang\textsuperscript{1}\thanks{\ \ indicates equal contribution.}},
 \textbf{Mingye Li\textsuperscript{2}$^*$\thanks{\ \ Work done during internship at TsinghuaNLP.}},
 \textbf{Xu Han\textsuperscript{1}\thanks{\ \ indicates corresponding authors.}},
 \textbf{Chaojun Xiao\textsuperscript{1$\ddag$}},
 \textbf{Weilin Zhao\textsuperscript{1}},\\
 \textbf{Ao Sun\textsuperscript{3}},
 \textbf{Hao Zhou\textsuperscript{4}},
 \textbf{Jie Zhou\textsuperscript{4}},
 \textbf{Zhiyuan Liu\textsuperscript{1}},
 \textbf{Maosong Sun\textsuperscript{1}}
\\
 \textsuperscript{1}{NLP Group, DCST, IAI, BNRIST, Tsinghua University, Beijing, China.}
 \\
\textsuperscript{2}{Department of CS\&T, Central South University, Changsha, China.}
\\
\textsuperscript{3}{BUPT, Beijing, China.}
\textsuperscript{4}{Pattern Recognition Center, WeChat AI, Tencent Inc.}
\\
{\tt huang-yx21@mails.tsinghua.edu.cn, lmy2004@csu.edu.cn,}\\ 
{\tt han-xu@tsinghua.edu.cn, xiaocj20@mails.tsinghua.edu.cn}
}

\begin{document}
\maketitle
\begin{abstract}
While long-context inference is crucial for advancing large language model (LLM) applications, its prefill speed remains a significant bottleneck. 
Current approaches, including sequence parallelism strategies and compute reduction through approximate attention mechanisms, still fall short of delivering optimal inference efficiency.
This hinders scaling the inputs to longer sequences and processing long-context queries in a timely manner.
To address this, we introduce \name, an efficient long-context inference framework that leverages multi-host approximate attention to enhance prefill speed by reducing compute and enhancing parallelism simultaneously. 
\name~introduces a communication mechanism for essential key-value pairs within a sequence parallelism framework, enabling a faster inference speed while maintaining task performance.
We implement \name~by incorporating a tailored \textsc{FlashAttn} kernel alongside optimized distribution strategies, supporting diverse models and parallelism configurations.
\name~achieves speedups of up to 9.2$\times$, 4.2$\times$, and 1.6$\times$ compared with \textsc{FlashAttn}, \textsc{RingAttn}, and \textsc{StarAttn}, respectively, without any observable task performance degradation.
We provide the implementation and experiment code of \name~in \url{https://github.com/thunlp/APB}.
\end{abstract}

\section{Introduction}
\label{sec:intro}

Large language models (LLMs)~\citep{gpt4o, claude3, deepseek-v3} have demonstrated unprecedented proficiencies, pushing the boundaries of artificial intelligence research and practical applications.
Recent advancements are not only transforming usage paradigms but also empowering intelligent systems such as LLM-based agents~\citep{li2024review, qin2023tool,zhao2024longagent}, robotics~\citep{zeng2023large, kim2024survey}, and prompting methodologies~\citep{chu2023survey,sahoo2024systematic}.
These systems often rely on extended context inference. 
To address the growing demand for longer inputs, contemporary foundation models have been increasingly designed to support extended context lengths.
For instance, \texttt{Llama-3.1}~\citep{dubey2024llama3} supports up to 128K tokens, \texttt{Claude-3.5}~\citep{claude3} extends input capacity to 200K tokens, and \texttt{MiniMax-01}~\citep{li2025minimax} can even process input sequences up to 4M tokens.

\begin{figure}[t]
\begin{center}
\includegraphics[width=0.9\linewidth]{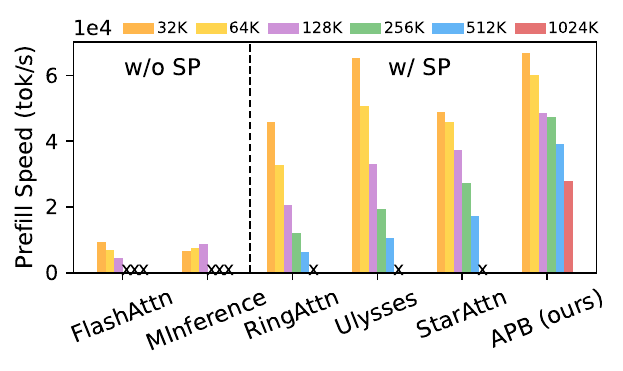}
\end{center}
\vspace{-18pt}
\caption{The prefill speed of methods with and without sequence parallelism when processing different input lengths. ``\texttt{SP}'' indicates sequence parallelism. ``\texttt{x}'' represents that the setting triggers out-of-memory error.}
\vspace{-12pt}
\label{fig:pt}
\end{figure}

As context lengths grow, the quadratic computational cost of attention makes single-GPU inference both infeasible and inefficient for LLMs.
To address this, various optimizations aim to \textit{enhance parallelism} or \textit{reduce compute}.
Sequence parallelism~\citep{li2021sequence}, which aims to enhance parallelism, partitions the sequence across devices (termed as \textit{hosts}) and significantly improves the prefill speed, especially for extremely long inputs (Figure~\ref{fig:pt}). 
However, the overall computations remain unchanged to ensure the accuracy of the attention results.
On the other hand, approximate attention mechanisms~\citep{zhang2024h2o, li2024snapkv, jiang2024minference}, which compute only those elements selected from the attention matrix, accelerate inference by reducing compute but face scalability challenges and performance degradation when processing longer inputs.
To this end, designing an approximate attention mechanism that fits in sequence parallelism frameworks offers a promising way to further enhance efficiency, particularly in accelerating long-context prefill.
However, \textit{designing such systems demands system and algorithm optimizations to address the key challenges.}

\underline{\textbf{Challenge 1:}} \textit{Localized Attention Pruning.}
Existing widely-used approximate attention mechanisms, such as \textsc{H$_2$O}~\citep{zhang2024h2o} and \textsc{SnapKV}~\citep{li2024snapkv}, typically depend on full sequence information, such as the attention scores computed over the entire sequence, to prune the redundant compute of the attention scores. 
This requirement directly conflicts with the distributed architecture of sequence parallelism, where individual hosts only maintain partial context visibility without heavy host-to-host communication.

\underline{\textbf{Challenge 2:}} \textit{Multi-host Scalability.}
Traditional sequence parallelism approaches face inherent scalability limitations due to model-architectural constraints and performance degradation risks. While sequence parallelism with attention head splitting for computation~\citep{jacobs2023deepspeed} offers substantial throughput improvements, its scalability remains fundamentally bounded by the fixed number of attention heads. Existing solutions that simply combine approximate attention with sequence parallelism, such as \textsc{StarAttn}~\citep{acharya2024star}, suffer from progressive performance degradation when the number of hosts increases, as these solutions merely extend~\citet{xiao2023efficient} to multiple hosts with a large proportion of invisible context.

To address these challenges, we propose \name{}, a distributed inference framework designed to leverage approximate attention to reduce redundant compute and communication overhead. For \textit{localized attention pruning}, \name{} introduces a local key-value (KV) cache compression technique that operates independently on each host, eliminating the need for a global sequence view to prune redundant attention compute. For \textit{multi-host scalability}, \name{} ensures that each host processes all attention heads within its local context and selectively shares compressed critical context across hosts. This design enables \name{} to maintain stable model performance even as the number of hosts scales up.

We implement \name{} using a customized \textsc{Flash-Attn}~\cite{dao2023flashattention2} kernel and an optimized distribution framework, enabling efficient scaling across diverse sequence lengths and multiple hosts. Comprehensive evaluations demonstrate that \name{} achieves an excellent trade-off between inference speed and model performance across a variety of tasks. Additionally, \name{} is compatible with various model sizes and distribution settings, making it a robust solution for scalable distributed inference. \name{} achieves speedups of up to 9.2$\times$, 4.2$\times$, and 1.6$\times$ compared with \textsc{FlashAttn}, \textsc{RingAttn}~\cite{li2021sequence}, and \textsc{StarAttn}, respectively, without any observable performance degradation.

\vspace{-2pt}
\section{Related Works}
\vspace{-2pt}
\label{sec:related}

Existing approaches to the efficient long-context inference of LLMs focus on system and algorithm optimizations. Please refer to the surveys~\cite{zhao2023survey, wan2023efficient} for more about LLMs.

\textbf{System Optimizations.}  
The efficiency of long-context inference can be enhanced by leveraging hardware architectures and preserving accurate LLM computations. 
Hardware-aware algorithms, such as \textsc{FlashAttn}~\citep{dao2022flashattention, dao2023flashattention2, shah2024flashattention}, utilize matrix tiling to optimize GPU memory usage and boost inference speed. 
Additionally, efficient parallelism methods support longer sequences with a higher processing speed. 
\textsc{RingAttn}~\citep{li2021sequence} splits long sequences across multiple hosts using ring-style communication to preserve accurate attention computation, while \textsc{Ulysses}~\citep{jacobs2023deepspeed} distributes attention heads across hosts to reduce communication overhead. 
Other parallelism strategies~\citep{narayanan2021efficient, huang2019gpipe, rajbhandari2020zero, sun2024seq1f1b, deepseek-v3} enable longer sequences on larger models, and the mixture of various distribution strategies~\citep{ren2021zero, ao2024burstattention} can further enhance long-context efficiency. 
However, most parallelism approaches are tailored for training and lack optimization for inference. 
Offloading-based methods~\cite{xiao2024infllm, sun2024shadowkv, lee2024infinigen, chen2024magicpig} leverage hierarchical memory systems to reduce hardware requirements by storing redundant KV cache in CPU memory and recalling only a small part to GPU memory.

\textbf{Algorithm Optimizations.}  
The burden of long-context inference can also be mitigated through algorithm optimizations. 
KV cache-centric optimization reduces the size of the KV cache, enabling faster decoding speed and lower GPU memory usage. 
Cache eviction methods~\cite{zhang2024h2o, li2024snapkv, yao2024sirllm, huang2024locret} discard irrelevant or unimportant KV cache, alleviating the memory bottlenecks. 
Quantization techniques~\cite{liu2024kivi, hooper2024kvquant, zhang2024kv,he2024zipcache} decrease the memory footprint by utilizing low-bit representations to store the KV cache. 
Merging methods~\cite{cai2024lococo, wang2024model, zhang2024simlayerkv, zhangcam} consolidate redundant KV cache units across sequences or layers to alleviate storage overheads. 
Sparse mechanisms~\cite{zaheer2020big, beltagy2020longformer, lou2024sparser} decrease inference latency by reducing the attention computational load. Specifically, approaches like \textsc{MInference}~\cite{jiang2024minference} and \textsc{FastGen}~\cite{ge2023model} assign specific patterns to attention heads, accelerating inference by computing only elements selected from the attention matrix. 
Moreover, algorithm optimizations can complement system optimizations.
\textsc{StarAttn}~\cite{acharya2024star} and \textsc{APE}~\cite{yang2025ape} linearize attention complexity by dividing context into parallelized blocks, but they struggle with tasks that require inter-context dependencies.
More details on these algorithm optimizations are elaborated in~\citet{li2024scbench} and~\citet{luohe2024keep}.
Apart from KV cache optimizations on Transformer-based~\cite{vaswani2017attention} LLMs, approaches altering backbone architectures can also enhance long-context inference efficiency.
Typically, RNN-based models~\citep{gu2023mamba,dao2024transformers} and hybrid models~\citep{lieber2024jamba,dong2024hymba,li2025minimax} reduce the computation complexity of long-context inference.

\section{Methodology}

\begin{figure*}[t]
\begin{center}
\includegraphics[width=\linewidth]{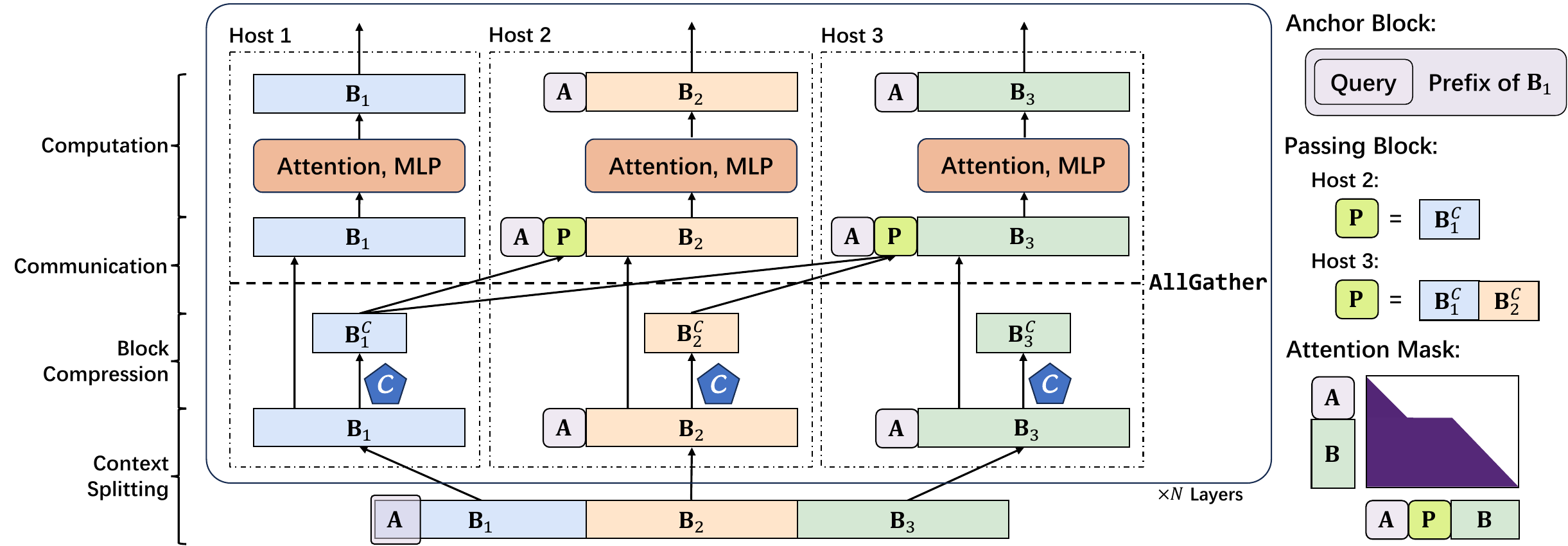}
\end{center}
\vspace{-5pt}
\caption{The framework of \name. The input document $d$ is split into blocks $\cblock{\mathbf{B}}_1, \cblock{\mathbf{B}}_2, \cblock{\mathbf{B}}_3$ and distributed across 3 hosts. The anchor block is denoted as ``$\canchor{\mathbf{A}}$'', the passing block as ``$\cpass{\mathbf{P}}$'', and the compressor as ``$\mathcal{C}$''. Each block is first prepended with an anchor block. When calculating attention, the context block $\cblock{\mathbf{B}}$ is compressed into $\cpass{\mathbf{B}}^C$ using the compressor $\mathcal{C}$. Subsequently, the passing block is constructed after an \texttt{AllGather} communication. Finally, attention is performed using a modified attention mask. Passing blocks are discarded after the attention computation.}
\vspace{-10pt}
\label{fig:framework}
\end{figure*}

\subsection{Preliminaries}

We primarily introduce the Transformer architecture and the prefill-decoding pattern.

\textbf{Transformers. }
Transformer-based LLMs take a sequence $\{t_1, t_2, \cdots, t_n\}$ as input.
We denote the hidden state of the token $t_k$ at the $i$-th layer as $\mathbf{H}^{(i)}[k]$, where $\mathbf{H}^{(0)}$ represents the embeddings of the input sequence.
Assuming the model has $L$ layers, each layer consists of an attention block and a feedforward network (FFN) block.
For each attention head in an attention block, the attention score at layer $i \in \{1, 2, \cdots, L\}$ is calculated as follows, where $\mathbf{M}$ represents the attention mask.
\begin{equation}
\small
    \mathbf{A}^{(i)} = \texttt{softmax}\left(\mathbf{M}\odot \frac{\mathbf{Q}^{(i)}{\mathbf{K}^{(i)\top}}}{\sqrt{d_m}}\right)\cdot \mathbf{V}^{(i)},
\end{equation}
where $\mathbf{Q}^{(i)} = \mathbf{H}^{(i-1)} \mathbf{W}_Q^{(i)}$, $\mathbf{K}^{(i)} = \mathbf{H}^{(i-1)} \mathbf{W}_K^{(i)}$, and $\mathbf{V}^{(i)} = \mathbf{H}^{(i-1)} \mathbf{W}_V^{(i)}$. The hidden size of each head is denoted by $d_m$.
Then, the attention scores are passed through the FFN: $\mathbf{H}^{(i)} = \texttt{FFN}(\mathbf{A}^{(i)})$.

\textbf{Prefill and Decoding. } 
The procedure of an LLM $\mathcal{M}$ processing an input can be divided into two stages: prefill and decoding.
During the prefill stage, the model computes the KV cache for the input sequence $\{t_1, t_2, \cdots, t_n\}$ and predicts the first answer token $t_{n+1}$.
This process can be formalized as $\left(t_{n+1}; \mathbf{K}[1:n], \mathbf{V}[1:n]\right) = \mathcal{M}\left(t_1, \cdots, t_n\right)$. After the prefill stage, the model decodes one token at a time using the KV cache of previous tokens, which can be written as $\left(t_{k+1}; \mathbf{K}[k], \mathbf{V}[k]\right) = \mathcal{M}\left(t_{k}; \mathbf{K}[1:k-1], \mathbf{V}[1:k-1]\right)$. The complexities of the prefill stage and a single step of decoding are $\mathcal{O}(n^2)$ and $\mathcal{O}(n)$, respectively.

\subsection{Framework}

We adopt sequence parallelism with approximate attention to accelerate the prefill of long-context inputs. The framework of \name~is illustrated in Figure~\ref{fig:framework}. 
The inference of \name~can be split into 4 stages: context splitting, block compression, communication, and computation. 

\textbf{Context Splitting.}
Given a long-context input $t = \{t_1, t_2, \cdots, t_{n}\}$, we first split the input sequence into a document $d = \{d_1, \cdots, d_{l_d}\}$ and a query $q = \{q_1, \cdots, q_{l_q}\}$, i.e., $t = \{d, q\}$ and $n = l_d + l_q$. 
Since $l_d \gg l_q$, we primarily focus on optimizing the prefill speed of the document.
We apply accurate attention with online softmax 
for processing the query and decoding the answer.
The document is evenly split across all hosts, with an \textit{anchor block} prepended at the start of each local context block on every host to preserve its visibility to the document's initial part.

\textbf{Block Compression.}
To reduce the attention compute and communication overhead, we first perform KV cache compression using a \textit{compressor} $\mathcal{C}$ on each host to shorten the block length and only retain the essential KV cache units for inter-host communication in the next step.
These compressed blocks are crucial for capturing long-distance semantic dependency and maintaining task performance by reducing attention errors.

\textbf{Communication.}
We design a specialized communication pattern to gather the compressed context blocks on each host for the awareness of previous essential context. 
After the communication, we form the \textit{passing blocks} to encapsulate the essential KV cache units passed from the previous hosts.

\textbf{Computation.}
Finally, each host obtains a context layout containing the local context block, an anchor block, and a passing block. 
We perform attention using a modified attention mask, executed with a specialized \textsc{FlashAttn}~\citep{dao2023flashattention2} kernel.

Overall, \name~is designed to communicate and compute only the most critical context by aggressively compressing the KV cache within a sequence parallelism framework. It introduces an algorithm-aware, distributed system-level optimization specifically tailored to its unique communication pattern and approximate attention mechanism.

\subsection{Context Splitting}

At the beginning of \name~inference, we first split the document $d$ across $H$ hosts, with each host holding a partial sequence with length $l_b = l_d / H$, denoted as blocks $\cblock{\mathbf{B}}_1, \cblock{\mathbf{B}}_2, \cdots, \cblock{\mathbf{B}}_H$. 
Each block is prepended with an anchor block at the front. 
Inspired by \textsc{StarAttn} ~\cite{acharya2024star}, we adopt the anchor block $\canchor{\mathbf{A}}$ containing the first $l_a$ tokens $d_1, \cdots, d_{l_a}$ of the input sequence.
We also embed the query $q$ in the front of the anchor block, to provide more query-related information for the compressors to identify the essential cache units. 
Therefore, the anchor block can be formally written as $\canchor{\mathbf{A}} = \{\canchor{q_1, \cdots, q_{l_q},~d_1, \cdots, d_{l_a}}\}$. The positions assigned to the tokens in anchor blocks are the starting positions, i.e. $0, 1, \cdots, l_q+l_a-1$.
Notably, we apply much smaller anchor blocks compared to \textsc{StarAttn}, where $l_a=\frac{1}{4}l_b$ or $l_a=\frac{1}{8}l_b$ in \name~but $l_a=l_b$ in \textsc{StarAttn}. 
Formally, each host except the first one holds a sequence of 
$
    \{\canchor{\mathbf{A}}, \cblock{\mathbf{B}}_h\} = 
    \{\canchor{q_1, \cdots, q_{l_q},~d_1, \cdots, d_{l_a}};~\cblock{ d_{(h-1)l_b+1}, \cdots, d_{hl_b}}\}
$.
The first host only holds $\cblock{\mathbf{B}}_1$ without anchor block.

\subsection{Block Compression}

Before starting the attention computation on the $h$-th host, we compress the KV cache of $\cblock{\mathbf{B}}_h$ into $\mathbf{\cpass{B}}^C_h$ using the compressors $\mathcal{C}$. 
As calculating full attention scores or full KV cache on a single host is infeasible, existing methods like \textsc{H$_2$O}~\cite{zhang2024h2o} and \textsc{SnapKV}~\cite{li2024snapkv} are incompatible in this scenario. 
The need to identify important KV cache units without a global view aligns with the design of \textsc{Locret}~\cite{huang2024locret}, where importance scores are based on each token's query and KV cache. 
We implement the compressors $\mathcal{C}$ as \textsc{Locret}'s retaining heads $\mathcal{R}$, which are small MLPs trained on additional long-context SFT data. 
These MLPs take $[\mathbf{Q}, \mathbf{K}, \mathbf{V}]$ as input and output importance scores for the corresponding cache units, reflecting the influence towards the calculation of subsequent KV cache. 
Formally, the compressed context block on host $h$ with passing length $l_p$ is defined as
$
\mathbf{\cpass{B}}^C_h = \left(\cpass{\mathbf{K}}_h^C, \cpass{\mathbf{V}}_h^C\right) = \left(\left\{\cblock{\mathbf{k}}_h[i_1], \cdots, \cblock{\mathbf{k}}_h[i_{l_p}]\right\}, ~ \left\{\cblock{\mathbf{v}}_h[i_1], \cdots, \cblock{\mathbf{v}}_h[i_{l_p}]\right\}\right),
$
where $\{s_{i_1}, \cdots, s_{i_{l_p}}\} = \texttt{Top-}l_p(s_1, \cdots, s_{l_b})$ and $s_1, \cdots, s_{l_b} = \mathcal{R}([\cblock{\mathbf{Q}}_h, \cblock{\mathbf{K}}_h, \cblock{\mathbf{V}}_h])$.

\begin{figure*}[t]
\begin{center}
\includegraphics[width=0.9\linewidth]{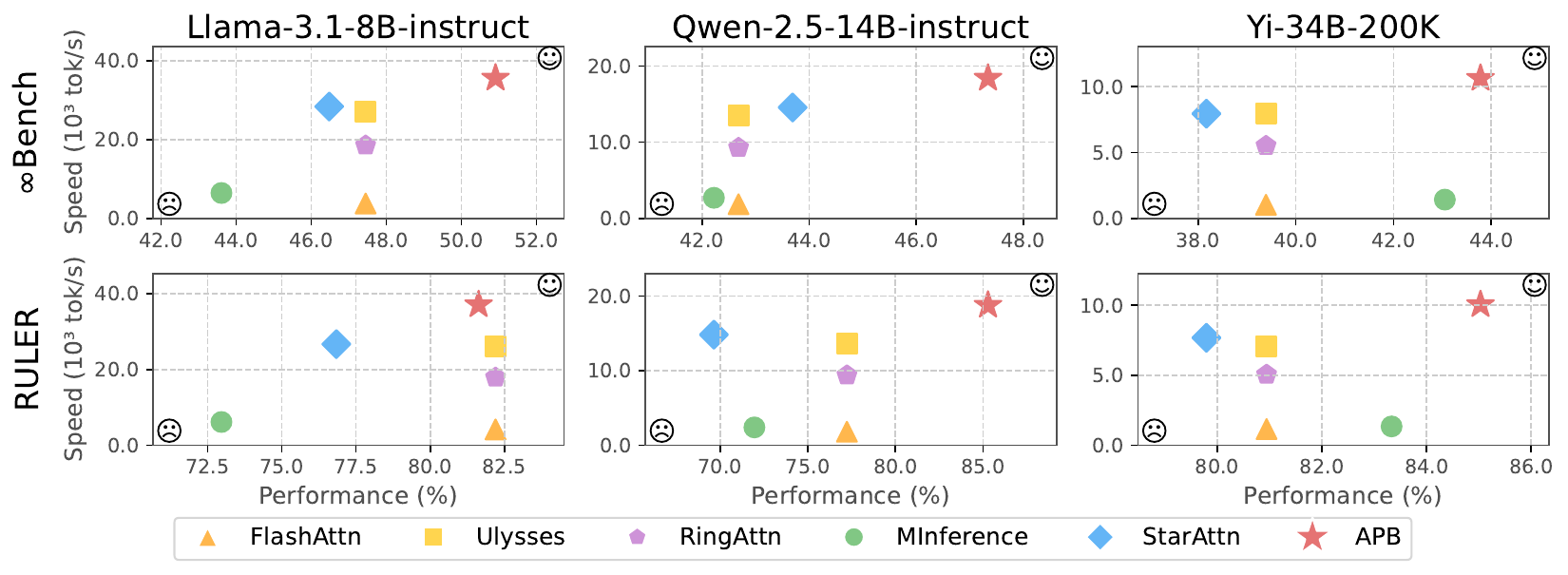}
\end{center}
\vspace{-5pt}
\caption{The inference speed and model performance of \name~and all the baselines. The top-right direction represents the optimal tradeoff between speed and performance. \name~achieves the best tradeoff of the two metrics. \textsc{FlashAttn}, \textsc{RingAttn}, and \textsc{Ulysses} share the same performance as they are all \textsc{FullAttn} methods.}
\vspace{-10pt}
\label{fig:acc-speed}
\end{figure*}

\subsection{Communication}
To acquire all the compressed context blocks sent from each host after block compression, we apply an \texttt{AllGather} communication on the compressed KV cache across all the hosts by $\left(\cpass{\mathbf{K}}_{[1:H]}^C,~ \cpass{\mathbf{V}}_{[1:H]}^C\right) = \texttt{AllGather}\left(\cpass{\mathbf{K}}_h^C, \cpass{\mathbf{V}}_h^C\right)$.
Subsequently, the passing block is constructed by concatenating all compressed context blocks from previous hosts, i.e.
$\cpass{\mathbf{P}}_h = \left(\cpass{\mathbf{K}}_{p}^C, ~\cpass{\mathbf{V}}_{p}^C\right) = \left(\cpass{\mathbf{K}}_{[1:h-1]}^C, ~\cpass{\mathbf{V}}_{[1:h-1]}^C\right)$. 
We simply ignore the compressed context blocks sent by subsequent hosts.

\subsection{Computation}

After the communication and construction of the passing blocks, each host acquires the context layout required for attention computation. 
We place the passing block between the anchor block and the local context block on each host to perform the attention as Equation~\ref{eq:apb}.
The passing blocks are discarded after the attention calculation and do not participate in the \texttt{FFN} calculation.
$\mathbf{M}'$ is the modified attention mask. We implement the attention calculation in a modified \textsc{FlashAttn}~\cite{dao2023flashattention2} kernel with only the attention mask changed.

We present the complete inference procedure of \name~in Algorithm~\ref{alg:infer}. Running approximate attention of the document $d$ is detailed in Algorithm~\ref{alg:apb_prefill}, while the accurate attention with online softmax is presented in Algorithm~\ref{alg:apb_decode}.
\begin{equation}
\small
\label{eq:apb}
\begin{aligned}
    \mathbf{Q}^{(i)} &= \left[\canchor{\mathbf{Q}}_a^{(i)}, \cblock{\mathbf{Q}}_h^{(i)}\right], \\
    \mathbf{K}^{(i)} &= \left[\canchor{\mathbf{K}}_a^{(i)},\cpass{\mathbf{K}}_p^{C},\cblock{\mathbf{K}}_h^{(i)}\right], \\
    \mathbf{V}^{(i)} &= \left[\canchor{\mathbf{V}}_a^{(i)},\cpass{\mathbf{V}}_p^{C},\cblock{\mathbf{V}}_h^{(i)}\right], \\
     \left[\canchor{\mathbf{A}}_a^{(i)}, \cblock{\mathbf{A}}_h^{(i)}\right] &= \texttt{softmax}\left(\mathbf{M}'\odot \frac{\mathbf{Q}^{(i)}{\mathbf{K}^{(i)\top}}}{\sqrt{d_m}}\right)\cdot \mathbf{V}^{(i)}, \\
     \left[\canchor{\mathbf{H}}_a^{(i)}, \cblock{\mathbf{H}}_h^{(i)}\right] &= \texttt{FFN}\left(\left[\canchor{\mathbf{A}}_a^{(i)}, \cblock{\mathbf{A}}_h^{(i)}\right]\right).
\end{aligned}
\end{equation}

\begin{table*}[ht!]
\small
\centering
\scalebox{0.88}{
\begin{tabular}{l|cccccccccc|c}
\toprule
Method & R.PassKey & R.Number & R.KV &  E.Sum & E.QA & E.MC & E.Dia & Z.QA & C.Debug & M.Find & Avg.\\
\midrule
\multicolumn{12}{c}{\texttt{Llama-3.1-8B-instruct}} \\ \midrule
\textsc{FullAttn} & 
100.00 & 99.49 & 51.00 & 30.59 & 29.04 & 63.76 & 11.00 & 36.18 & 24.62 & 28.82 & 47.45
\\ \midrule
\textsc{MInference} &
98.47 & 98.81 & 17.40& 30.06 & 26.42 & 55.46 & 12.50 & 36.51 & 28.17 & \textbf{32.29} & 43.61
\\
\textsc{StarAttn} &
\textbf{100.00} & \textbf{98.98} & 40.60& 30.55 & \textbf{30.66} & 61.57 & \textbf{15.00} & 36.26 & 26.65 & 24.57 & 46.48
\\
\rowcolor{pink!20}
\textbf{\name} &
\textbf{100.00} & 98.81 & \textbf{81.8} & \textbf{30.63} & 29.25 & \textbf{63.32} & 11.00 & \textbf{36.99} & \textbf{30.46} & 26.86 & \textbf{50.91}
\\
\midrule
\multicolumn{12}{c}{\texttt{Qwen-2.5-14B-instruct}} \\
\midrule
\textsc{FullAttn} & 
100.00& 100.00& 17.80& 27.80& 10.40& 52.84 & 28.00& 10.21 & 38.07 & 42.57 & 42.68
\\ \midrule
\textsc{MInference} &
\textbf{100.00}& \textbf{100.00}& 5.20& 25.63 & 12.04 & 61.57 & 16.50& 11.06 & 41.62 & \textbf{48.57} & 42.22
\\
\textsc{StarAttn} &
\textbf{100.00}& \textbf{100.00}& 18.00& \textbf{27.48} & 12.44 & 62.88 & 23.50& 11.03 & \textbf{43.40}& 37.71 & 43.69
\\
\rowcolor{pink!20}
\textbf{\name} &
\textbf{100.00}& \textbf{100.00}& \textbf{62.20}& 26.51 & \textbf{12.74} & \textbf{66.38} & \textbf{25.00}& \textbf{11.22} & 37.31 & 35.71 & \textbf{47.34}
\\
\midrule
\multicolumn{12}{c}{\texttt{Yi-34B-200K}} \\
\midrule
\textsc{FullAttn} & 
100.00 & 100.00 & 49.00 & 5.83 & 17.57 & 47.60 & 2.00 & 18.77 & 25.13 & 28.00 & 39.39
\\ \midrule
\textsc{MInference} &
\textbf{100.00} & \textbf{100.00} & 52.40 & \textbf{7.19} & \textbf{20.07} & \textbf{63.32} & \textbf{2.50} & 25.40 & \textbf{28.17} & \textbf{31.40} & 43.05
\\
\textsc{StarAttn} &
\textbf{100.00} & \textbf{100.00} & 24.60 & 4.38 & 19.71 & 60.26 & 0.00 & 23.40 & 28.17 & 21.14 & 38.17 
\\
\rowcolor{pink!20}
\textbf{\name} &
\textbf{100.00} & \textbf{100.00} & \textbf{65.20} & 5.96 & 19.81 & 62.45 & 1.00 & \textbf{25.42} & 27.92 & 30.00 & \textbf{43.78}
\\

\bottomrule
\end{tabular}
}
\vspace{-5pt}
\caption{The results of \name~compared with all the baselines on $\infty$Bench, where higher score represents better performance. ``Avg.'' represents the average score. The highest score in each column is marked in \textbf{bold}. \textsc{FullAttn} represents \textsc{FlashAttn}, \textsc{RingAttn}, and \textsc{Ulysses}, as their computational results remain unchanged.}
\label{tab:results-infinitebench}
\end{table*}

\begin{table*}[t!]
\vspace{-5pt}
\small

\centering
\scalebox{0.803}{
\begin{tabular}{l|ccccccccccccc|c}
\toprule
Method & SG1 & SG2 & SG3 & MK1 & MK2 & MK3 & MV & MQ & VT & CWE & FWE & QA1 & QA2 & Avg.\\
\midrule
\multicolumn{15}{c}{\texttt{Llama-3.1-8B-instruct}} \\ \midrule
\textsc{FullAttn} & 
99.40 & 99.80 & 99.60 & 98.20 & 87.60 & 67.00 & 94.65 & 98.00 & 60.98 & 71.40 & 72.20 & 78.20 & 41.6 & 82.20
\\ \midrule
\textsc{MInference} &
\textbf{100.00} & 98.60 & 99.00 & 95.40 & 58.20 & 23.80 & 84.35 & 95.70& \textbf{66.40} & 45.94 & 74.67 & 67.80 & 38.80 & 72.97
\\
\textsc{StarAttn} &
\textbf{100.00} & 99.60 & 99.60 & \textbf{95.80} & 73.60 & 53.00 & 72.80 & 94.45 & 59.40 & \textbf{65.72} & 76.53 & 67.40 & 41.00 & 76.84 
\\
\rowcolor{pink!20}
\textbf{\name} &
\textbf{100.00} & \textbf{100.00} & \textbf{99.80} & 85.60 & \textbf{91.00} & \textbf{89.00} & \textbf{95.05} & \textbf{96.40} & 51.96 & 63.82 & \textbf{77.33} & \textbf{70.00} & \textbf{41.20} & \textbf{81.63}
\\
\midrule
\multicolumn{15}{c}{\texttt{Qwen-2.5-14B-instruct}} \\
\midrule
\textsc{FullAttn} & 
\textbf{100.00} & 99.20 & 99.80 & 94.20 & 47.80 & 27.20 & 75.10 & 94.60 & 89.52 & 93.88 & 76.13 & 63.20 & 43.40 & 77.23 
\\ \midrule
\textsc{MInference} &
\textbf{100.00} & 99.60 & \textbf{100.00} & 89.60 & 32.60 & 6.94 & 61.25 & \textbf{93.29} & 89.44 & 84.36 & 77.33 & 56.60 & 44.40 & 71.95
\\
\textsc{StarAttn} &
\textbf{100.00} & 99.20 & 97.40 & 80.00 & 38.00 & 20.20 & 48.75 & 77.10 & 82.44 & 93.86 & \textbf{77.53} & 52.80 & 38.00 & 69.64
\\
\rowcolor{pink!20}
\textbf{\name} &
\textbf{100.00} & \textbf{100.00} & 99.80 & \textbf{93.00} & \textbf{87.20} & \textbf{85.80} & \textbf{66.55} & 93.00 & \textbf{96.40} & \textbf{94.94} & 76.33 & \textbf{60.00} & \textbf{55.60} & \textbf{85.28}
\\
\midrule
\multicolumn{15}{c}{\texttt{Yi-34B-200K}} \\
\midrule
\textsc{FullAttn} & 
100.00 & 100.00 & 99.60 & 95.20 & 76.00 & 55.40 & 92.10 & 97.05 & 85.56 & 51.84 & 84.27 & 65.20 & 50.00 & 80.94
\\ \midrule
\textsc{MInference} &
\textbf{100.00} & \textbf{100.00} & \textbf{100.00} & \textbf{96.00} & 87.00 & 59.60 & \textbf{90.15} & 96.05 & 73.68 & 79.20 & \textbf{84.87} & \textbf{67.40} & 49.40 & 83.33
\\
\textsc{StarAttn} &
\textbf{100.00} & \textbf{100.00} & 99.60 & 93.20 & 78.20 & 48.00 & 77.05 & 88.35 & \textbf{83.96} & 80.58 & 78.33 & 61.20 & 48.80 & 79.79
\\
\rowcolor{pink!20}
\textbf{\name} &
\textbf{100.00} & \textbf{100.00} & \textbf{100.00} & 92.80 & \textbf{88.00} & \textbf{92.60} & 86.05 & \textbf{96.45} & 60.48 & \textbf{88.46} & 84.53 & 63.80 & \textbf{52.20} & \textbf{85.03}
\\

\bottomrule
\end{tabular}
}
\vspace{-5pt}
\caption{The results of \name~compared with all the baselines on RULER, where higher score represents better performance. ``Avg.'' represents the average score. The highest score in each column is marked in \textbf{bold}. \textsc{FullAttn} represents \textsc{FlashAttn}, \textsc{RingAttn}, and \textsc{Ulysses}, as their computational results remain unchanged.
}
\label{tab:results-ruler}
\vspace{-5pt}
\end{table*}

\section{Experiments}

In this section, we present our benchmark results to evaluate \name,
aiming at addressing the following research questions:
\noindent\:(\underline{\textbf{Q1}}) Can \name~obtain higher end-to-end prefill speed without significant performance degradation? 
\noindent\:(\underline{\textbf{Q2}}) How effective and efficient is \name~on various context length?
\noindent\:(\underline{\textbf{Q3}}) Is the design of each component in \name~effective?

\subsection{Experimental Setup}

In this section, we briefly introduce experimental settings, and more details are in Appendix~\ref{appendix:hyper}.

\textbf{Benchmarks.} We evaluate \name~against selected baselines on two long-context evaluation benchmarks: $\infty$Bench~\citep{zhang2024infty} and RULER~\cite{hsieh2024ruler}. $\infty$Bench is a benchmark with an average context length exceeding 100K tokens, encompassing a mix of both synthetic and real-world tasks. Following \textsc{MInference}~\cite{jiang2024minference}, we utilize 10 tasks, excluding Math.Calc and Code.Run due to their difficulty, which leads to failures for all methods. RULER is a benchmark featuring a variety of synthetic tasks with a controllable context length. We evaluate all 13 tasks of the benchmark. We provide the mapping of task abbreviations to the original task name in Appendix~\ref{sec:abbre}. Additional experiments and results are placed in Appendix~\ref{sec:detailed_results}.

\textbf{Models.} For the end-to-end benchmark (\underline{\textbf{Q1}}), we test \name~with the baselines on the instruct versions of \texttt{Llama-3.1-8B}, \texttt{Qwen-2.5-14B}, and \texttt{Yi-34B-200K}. 
When benchmarking across various input lengths (\underline{\textbf{Q2}}), we utilize \texttt{Llama-3-8B-1M}\footnote{https://huggingface.co/gradientai/Llama-3-8B-Instruct-Gradient-1048k}, which supports up to 1M input tokens.

\textbf{Metrics.} When evaluating performance on the benchmark, we directly adopt the original performance metrics for each task. For speed measurement, we define inference speed as the total number of tokens in both the prefill and decoding stages, divided by the total inference time, which includes both the prefill and decoding stages.

\textbf{Baselines.} 
Existing methods for accelerating long-context inference can be categorized into four types based on their use of sequence parallelism and approximate attention mechanisms. We select \textsc{FlashAttn}, \textsc{Ulysses}, \textsc{RingAttn}, \textsc{MInference}, and \textsc{StarAttn} as our baselines. \textsc{FlashAttn} does not use sequence parallelism and maintains accurate attention computation. \textsc{Ulysses} and \textsc{RingAttn} are two major approaches for performing accurate attention with sequence parallelism, each with a different communication design. \textsc{MInference} is an approximate attention mechanism that accelerates long-context inference by applying sparse attention without sequence parallelism. 
\textsc{StarAttn} combines sequence parallelism and approximate attention by prepending an anchor block to each context block and removing all communication. 
Further details on the baselines can be found in Appendix~\ref{appendix:baselines}.

\begin{figure*}[t]
\captionsetup[subfigure]{
  labelformat=simple,
}
\begin{center}
\subfloat[Performance]{
    \includegraphics[width=0.32\linewidth]{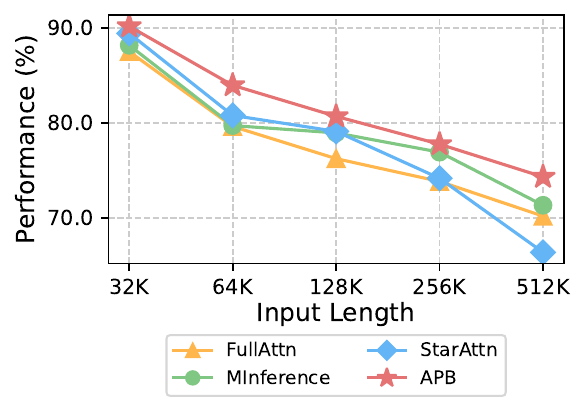}
    \vspace{-6pt}
    \label{fig:var-seqlen-a}
}
\subfloat[Speed]{
    \includegraphics[width=0.32\linewidth]{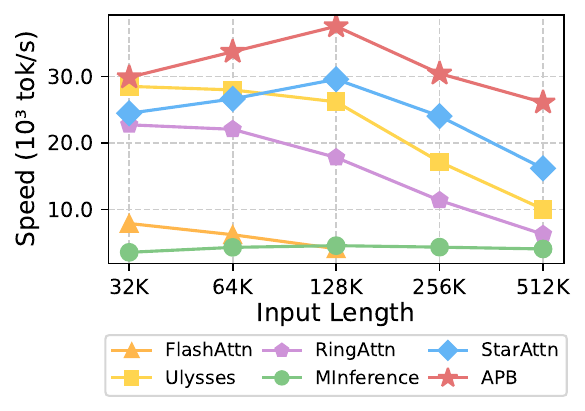}
    \vspace{-6pt}
    \label{fig:var-seqlen-b}
}
\subfloat[Compute]{
    \includegraphics[width=0.327\linewidth]{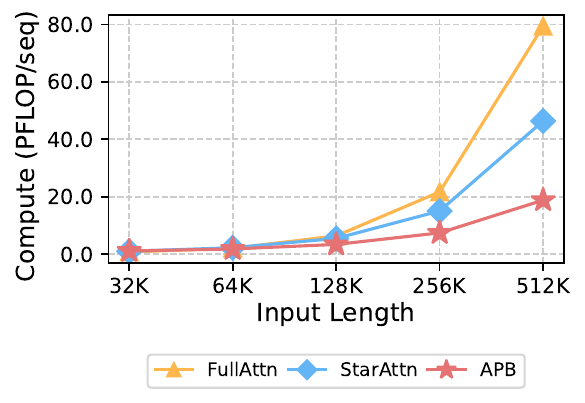}
    \vspace{-6pt}
    \label{fig:var-seqlen-c}
}
\end{center}
\vspace{-10pt}
\caption{The performance, speed, and the amount of compute of various methods under different input lengths. \name~consistently outperforms other methods with better performance, faster speed, and lower compute.}
\vspace{-10pt}
\label{fig:var-seqlen}
\end{figure*}

\subsection{End-to-End Results}
\label{sec:end-to-end}

To address \underline{\textbf{Q1}}, we conduct an end-to-end benchmark focusing on the overall performance and long-context processing speed by comparing \name~with all the baselines on both $\infty$Bench and RULER.

\name~demonstrates superior task performance in both real-world scenarios ($\infty$Bench in Table~\ref{tab:results-infinitebench}) and synthetic benchmarks (RULER in Table~\ref{tab:results-ruler}), compared to all the selected baselines.
In certain tasks, \name~even surpasses \textsc{FullAttn}.
In contrast, \textsc{MInference} and \textsc{StarAttn} exhibit noticeable performance degradation relative to \textsc{FullAttn}.
Notably, \name~significantly enhances performance in complex context retrieval tasks, such as R.KV in $\infty$Bench and MK3 in RULER. 
This improvement can be attributed to \name's ability to trim noisy context within each block by our compressor, thereby constructing cleaner passing blocks with reduced noise and enabling more accurate context retrieval.

In terms of inference speed, \name~also demonstrates significant advantages over existing methods. 
As illustrated in Figure~\ref{fig:acc-speed}, methods without sequence parallelism, such as \textsc{FlashAttn} and \textsc{MInference}, exhibit slower inference speeds. 
In contrast, sequence parallelism-based methods, including \textsc{RingAttn} and \textsc{Ulysses}, achieve speedups ranging from 3$\times$ to 10$\times$ compared to \textsc{FlashAttn}. 
By incorporating approximate attention, \textsc{StarAttn} achieves a faster inference speed than \textsc{RingAttn}, though its improvement over \textsc{Ulysses} remains limited. 
\name, on the other hand, 
delivers at least a 25\% speed improvement over \textsc{StarAttn} without any performance decay. 

Overall, \name~establishes the best speed-performance trade-off, outperforming all competing methods in both efficiency and effectiveness.
 
\subsection{Benchmarking Across Different Lengths}
\label{sec:varlen}

We evaluate \name~and baselines with input lengths ranging from 32K to 512K to address \underline{\textbf{Q2}}. 
All methods are tested on the RULER benchmark, where we report task performance, inference speed, and the compute under each input length in Figure~\ref{fig:var-seqlen}. 

As shown in Figure~\mbox{\ref{fig:var-seqlen}\hspace{-20pt}\subref{fig:var-seqlen-a}}, \name~consistently outperforms all the methods across all input lengths in terms of performance.
Additionally, \name~achieves the fastest inference speed among all methods, as depicted in Figure~\mbox{\ref{fig:var-seqlen}\hspace{-20pt}\subref{fig:var-seqlen-b}}.
The speed advantage of \name~remains humble for shorter input sequences.
However, as the input length increases, \name's advantage becomes more pronounced, demonstrating a substantial speedup over all competitors at a 512K input length.
Notably, both \textsc{StarAttn} and \name~show an increase in inference speed from 32K to 128K inputs, whereas other methods continue to slow down.
Since \textsc{StarAttn} and \name~both incorporate approximate attention mechanisms, the overall compute is reduced, delaying the point where compute becomes the dominant bottleneck instead of memory access.
Figure~\mbox{\ref{fig:var-seqlen}\hspace{-20pt}\subref{fig:var-seqlen-c}} further explains \name's superior speed, as it yields significantly less compute than \textsc{FullAttn} and \textsc{StarAttn}, especially for longer inputs. 
Further discussion on the compute of each method can be referred to Table~\ref{tab:flops}.

\begin{figure*}[t]
\begin{center}
\includegraphics[width=\linewidth]{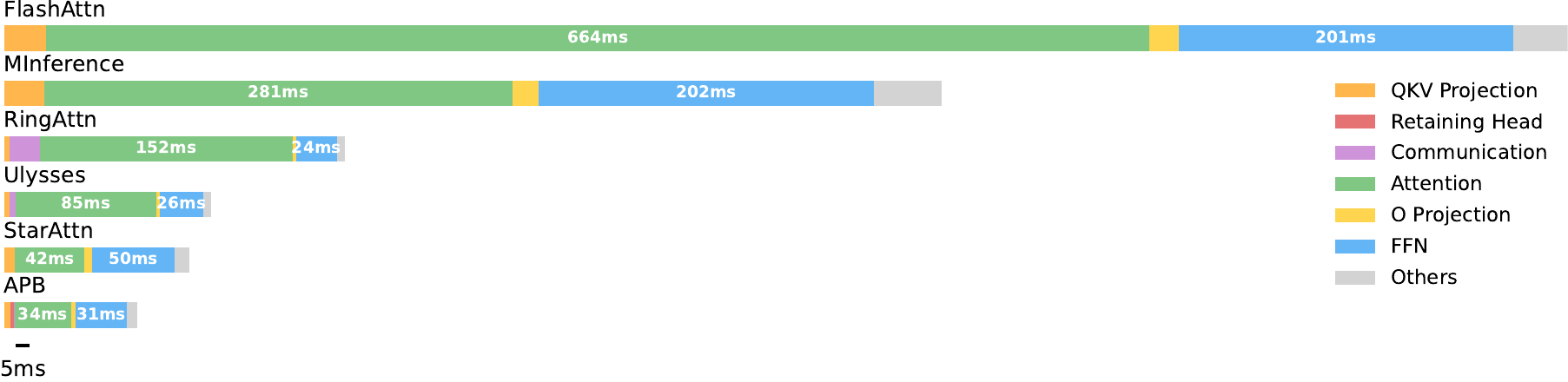}
\end{center}
\vspace{-5pt}
\caption{The wall-time breakdown of prefill for various methods on 128K context. }
\vspace{-10pt}
\label{fig:breakdown-tb}
\end{figure*}

\subsection{Ablation Studies}
\label{sec:abl}

Regarding \underline{\textbf{Q3}}, we conduct ablation studies and focus on the following key points.

\begin{table}[t]
\small
\center
\setlength{\tabcolsep}{12pt}
\scalebox{0.9}{
\begin{tabular}{l|cccc|c}
\toprule
No. & \canchor{\textbf{A}} & \cpass{\textbf{P}} & $\mathcal{C}$ & Q & E.MC \\
\midrule
0 & \ding{51} & \ding{51} & $\mathcal{R}$ & \ding{51} & \textbf{72.00} \\
\midrule
1 & \ding{51} & \ding{51} & $\mathcal{R}$ & \ding{55} & 68.00 \\
2 & \ding{51} & \ding{51} & Rd. & \ding{51} & 66.00 \\
3 & \ding{51} & \ding{51} & Rd. & \ding{55} & 66.00 \\
\midrule
4 & \ding{51} & \ding{55} & Rd. & \ding{51} & 64.00 \\
5 & \ding{51} & \ding{55} & Rd. & \ding{55} & 62.00 \\
\midrule
6 & \ding{55} & \ding{51} & $\mathcal{R}$ & \ding{55} & 28.00 \\
7 & \ding{55} & \ding{51} & Rd. & \ding{55} & 28.00 \\
\midrule
8 & \ding{55} & \ding{55} & Rd. & \ding{55} & 22.00 \\
\bottomrule
\end{tabular}}
\caption{Ablation studies of \name~on E.MC. ``\canchor{\textbf{A}}'' and ``\cpass{\textbf{P}}'' represent the anchor block and passing block. We implement the compressors $\mathcal{C}$ as the retaining heads $\mathcal{R}$ or random selectors ``Rd.''.
``Q'' indicates embedding the query in the anchor block. 
}
\label{tab:ablation}
\end{table}

\textbf{Each Component's Contribution.} We focus on four major components in the \name~framework: the anchor block, the passing block, selecting important KV pairs via retaining heads, and embedding the query at the beginning of the anchor block. We conduct an ablation study on these four components. For the anchor block and passing block, they can be removed in the corresponding settings. To ablate the implementation of the compressors $\mathcal{C}$, we compare retaining heads $\mathcal{R}$ with random selectors picking the same number of KV pairs. When the embedded query is removed, the anchor block contains only the beginning of the document. 

The ablation results in Table~\ref{tab:ablation} show that all four components are effective, as removing any of them leads to performance degradation (No. 1, 2, 4, 6). The anchor block is the most critical component, as its removal causes task failure (No. 6, 7). The passing block is also essential for good performance, as a performance drop of more than 8\% occurs in No. 4 and 5. The retaining heads are crucial for selecting important KV pairs to pass to subsequent hosts, as randomly selecting KV pairs results in poorer performance (No. 2). Embedding the query in the anchor block must be used in conjunction with the retaining heads, as activating these components separately does not improve performance (No. 1, 2, 3). Above all, when used together, they yield significantly better performance (No. 0).

\textbf{Wall-time Breakdown Analysis.} 
We begin by analyzing the prefill and decoding time, as detailed in Appendix~\ref{sec:break_pd}. Our findings indicate that prefill dominates the overall processing time, while decoding time is negligible in long-context tasks.

Next, we decompose the prefill time into seven components: the projection of QKV, the calculation of retaining heads, the communication time, the attention computation, the projection of O, the FFN computation, and others. The wall-time of each component is shown in Figure~\ref{fig:breakdown-tb}. Methods without sequence parallelism (\textsc{FlashAttn} and \textsc{MInference}) exhibit substantial inference time, whereas sequence parallelism proves effective in reducing the time of both attention and FFN computations. 

Compared to existing sequence parallelism methods, \name~achieves even lower attention time without incurring significant overhead for retaining heads and communication. When compared with \textsc{RingAttn} and \textsc{Ulysses}, although the introduction of the anchor blocks increases the compute cost for FFN, the additional time is smaller than the reduction in attention time, thus resulting in a faster inference speed. 
\name~outperforming \textsc{StarAttn} is due to smaller anchor blocks, whose overhead in the FFN time is significantly lower.

\textbf{Distributed Settings.} We evaluate the effectiveness of \name~across various distributed settings. As mentioned in \textsc{StarAttn}~\citep{acharya2024star}, when the context is split across more hosts, significant performance degradation occurs due to the increasing invisibility of the middle context. However, \name~addresses this issue by using passing blocks, allowing each host to share the most essential KV pairs with subsequent hosts. 

We test \name~and \textsc{StarAttn} across $\{2, 4, 6, 8\}$ hosts, and the results are shown in Table~\ref{tab:dist}. When the sequence is long (128K tokens), both \name~and \textsc{StarAttn} exhibit improved performance as the sequence parallelism size increases. 
With a shorter input length (32K tokens), the block size on each host is smaller, and \name~demonstrates consistently stronger and more stable performance across all settings, while \textsc{StarAttn} shows significant performance loss as the host count increases.

\begin{table}[h]
\small
\center
\scalebox{0.9}{
\begin{tabular}{ll|cccc}
\toprule
Length & Method & $H=2$ & $H=4$ & $H=6$ & $H=8$ \\
\midrule
\multirow{2}{*}{128K} & \name & 84.00 & 92.00 & 92.00 & 92.00 \\
& \textsc{StarAttn} & 82.00 & 84.00 & 86.00 & 88.00\\
\midrule
\multirow{2}{*}{32K} & \name & 92.00 & 92.00 & 94.00 & 94.00 \\
& \textsc{StarAttn} & 94.00 & 92.00 & 92.00 & 84.00 \\
\bottomrule
\end{tabular}}
\caption{The E.MC performance of \name~and \textsc{Star-Attn} tested under various distributed settings and different context length. ``$H$'' refers to the number of hosts.}

\label{tab:dist}
\end{table}

\textbf{Short-Context Performance.} \label{sec:short-context-main} While \name~ exhibits outstanding performance and inference speed on long-context benchmarks, its effectiveness and efficiency on short-context inputs (i.e., 4K tokens) are equally important. To evaluate this, we conduct experiments on RULER using 4K input lengths, reporting both task performance and inference speed. We compare \name~against \textsc{FlashAttn} using the \texttt{Llama-3-8B-instruct-1M} model, and present the results in Table~\ref{tab:shortlen-acc-abrev}.

The results demonstrate that \name~consistently achieves superior speed and performance on short-context inputs. Even in such scenarios, \name~outperforms \textsc{FullAttn} without any degradation in performance or slowdown in inference. Please refer to Appendix~\ref{sec:short-context} for more details.

\begin{table}[h]
\small

\centering
\scalebox{0.9}{
\begin{tabular}{l|c|c}
\toprule
Method &  RULER-4K & Speed(tok/s).\\
\midrule
\textsc{FlashAttn} &  94.54& 6247.58\\ 
\rowcolor{pink!20}
\textsc{\name~} & \textbf{94.89}& \textbf{6597.47} \\
\bottomrule
\end{tabular}
}

\caption{The task performance and inference speed of \name~and \textsc{FlashAttn} on RULER, where context length set to 4K tokens. We report the speed in  ``tok/s''.}
\label{tab:shortlen-acc-abrev}
\end{table}

\textbf{Orthogonality to Quantization.} \name~exhibits strong orthogonality and can be integrated with other compression methods, such as KV quantization. To demonstrate this, we tested \name~in combination with KV cache quantization techniques. We integrate the KV cache quantization method officially supported by Hugging Face Transformers\footnote{https://github.com/huggingface/transformers} into the \name~framework. This method is inspired by KIVI~\citep{liu2024kivi} and implemented using the HQQ backend\footnote{https://github.com/mobiusml/hqq}.
We evaluate the integration on the RULER benchmark using 50 entries per task with \texttt{LLaMA-3.1-8B-Instruct} as the backbone. The KV cache is quantized to 4 bits. For comparison, the results of \textsc{FullAttn}, \textsc{MInference}, \textsc{StarAttn}, and standalone \name~are taken from Table 2, which used 500 entries per task. The results are summarized in Table~\ref{tab:apb-quant}.

The results shows that \name~is compatible with KV quantization. In this setting, the stadalone \name~shows a minor performance drop compared with FullAttn, and \name+HQQ-4bits also shows a small performance degredation compared with standalone HQQ-4bits quantization. However, it is worth noticing that \name+HQQ-4bits can still ourperform the baselines (\textsc{MInference} and \textsc{StarAttn}), as the baselines do not use quantization. This setting is able to reduce the size of KV cache to 4$\times$ and enhance prefill speed at the meantime, showing a promising balance of task performance and inference efficiency.

\begin{table}[h]
\small
\centering
\scalebox{0.9}{
\begin{tabular}{l|cc}
\toprule
Method & Original (M) & Quantized (M-4bits) \\
\midrule
\textsc{FullAttn} & 82.20 & 81.73 \\
\textsc{MInference} & 72.97 & - \\
\textsc{StarAttn} & 76.84 & - \\
\rowcolor{pink!20}
\name & 81.63 & 79.13 \\
\bottomrule
\end{tabular}
}
\caption{Performance of \textsc{FullAttn}, \textsc{MInference}, \textsc{StarAttn}, \name, and their quantized variants on RULER-128K. ``M'' denotes the original method and ``M-4bits'' represents its 4-bit quantized version. Since \textsc{MInference} and \textsc{StarAttn} cause more degradation than quantization, we skip their quantized versions.}
\label{tab:apb-quant}
\end{table}

\textbf{Training Generalizability.} To demonstrate that the training of the retaining heads is robust and generalizable, we conduct an ablation study on their training data. Specifically, we train the retaining heads on datasets including LongAlpaca~\citep{chen2023longlora}, LongAlign, and AntiHaystack\footnote{https://huggingface.co/datasets/wenbopan/anti-haystack}, using \texttt{Llama-3.1-8B-instruct} as the backbone model. We then evaluate \name~with these trained retaining heads on RULER across 20 samples per task. Other settings are aligned to Table \ref{tab:results-ruler}. The results are presented in Table \ref{tab:apb-data}. 

The similar results across different training data indicate that the retaining heads in \name~are robust across different training datasets, demonstrating their generalizability. Consequently, various long-context training data recipes can be employed in the \name~framework. 

\begin{table}[h]
\small
\centering
\scalebox{0.9}{
\begin{tabular}{l|ccc}
\toprule
Training Data & LongAlign & LongAlpaca & AntiHaystack \\
\midrule
RULER-128K & 81.17 & 81.04 & 80.40 \\
\bottomrule
\end{tabular}
}
\caption{The task performance tested on RULER-128K of \name~trained on various datasets.}
\label{tab:apb-data}
\end{table}

\section{Conclusion}

We propose \name, a distributed long-context inference framework based on approximate attention, achieving speedups up to 9.2$\times$, 4.2$\times$, and 1.6$\times$ compared to \textsc{FlashAttn}, \textsc{RingAttn}, and \textsc{StarAttn}, respectively, without observable performance degradation. 
\name~constructs the passing blocks with minimal communication cost by sending only core KV pairs among hosts.
The experimental results on $\infty$Bench and RULER, using various models and sequence lengths, demonstrate that \name~delivers faster inference speeds while maintaining or exceeding performance. Furthermore, \name~is adaptable to diverse distribution configurations and models of varying sizes.
Next, we aim to accelerate the decoding process in \name, where the KV cache is distributed across different hosts.


\section*{Limitations}

As \name~is specifically optimized to minimize the prefill time for extremely long inputs, it is less effective for processing shorter inputs, particularly those under 32K tokens. 
When applying \name~to short inputs, the optimal distributed setting is to run the inference on a single host, as the computation can be effectively parallelized within a single GPU. 
In such cases, \name~falls back to vanilla \textsc{FlashAttn}.

\section*{Acknowledgments}

This work is supported by the National Key R\&D Program of China (No.2022ZD0116312) and a grant from the Guoqiang Institute, Tsinghua University. Yuxiang Huang is supported by Beijing National Science Foundation (No. QY24253).
We deeply thank the anonymous reviewers for their constructive reviews. 

\bibliography{acl_latex}

\begin{thebibliography}{60}
\providecommand{\natexlab}[1]{#1}

\bibitem[{Acharya et~al.(2024)Acharya, Jia, and Ginsburg}]{acharya2024star}
Shantanu Acharya, Fei Jia, and Boris Ginsburg. 2024.
\newblock Star {A}ttention: Efficient llm inference over long sequences.
\newblock \emph{ArXiv:2411.17116}.

\bibitem[{Anthropic(2024)}]{claude3}
Anthropic. 2024.
\newblock \href {https://www.anthropic.com/claude/sonnet} {Claude 3.5 {S}onnet}.

\bibitem[{Bai et~al.(2024)Bai, Lv, Zhang, He, Qi, Hou, Tang, Dong, and Li}]{bai2024longalign}
Yushi Bai, Xin Lv, Jiajie Zhang, Yuze He, Ji~Qi, Lei Hou, Jie Tang, Yuxiao Dong, and Juanzi Li. 2024.
\newblock Longalign: A recipe for long context alignment of large language models.
\newblock \emph{Proceedings of EMNLP}.

\bibitem[{Beltagy et~al.(2020)Beltagy, Peters, and Cohan}]{beltagy2020longformer}
Iz~Beltagy, Matthew~E Peters, and Arman Cohan. 2020.
\newblock Longformer: The long-document transformer.
\newblock \emph{ArXiv:2004.05150}.

\bibitem[{Cai et~al.(2024)Cai, Tian, Wang, and Chen}]{cai2024lococo}
Ruisi Cai, Yuandong Tian, Zhangyang Wang, and Beidi Chen. 2024.
\newblock Lo{C}o{C}o: Dropping in convolutions for long context compression.
\newblock \emph{Proceedings of ICML}.

\bibitem[{Chen et~al.(2024)Chen, Qian, Tang, Lai, Liu, Han, and Jia}]{chen2023longlora}
Yukang Chen, Shengju Qian, Haotian Tang, Xin Lai, Zhijian Liu, Song Han, and Jiaya Jia. 2024.
\newblock Longlora: Efficient fine-tuning of long-context large language models.
\newblock \emph{Proceedings of ICLR}.

\bibitem[{Chen et~al.(2025)Chen, Sadhukhan, Ye, Zhou, Zhang, Nolte, Tian, Douze, Bottou, Jia et~al.}]{chen2024magicpig}
Zhuoming Chen, Ranajoy Sadhukhan, Zihao Ye, Yang Zhou, Jianyu Zhang, Niklas Nolte, Yuandong Tian, Matthijs Douze, Leon Bottou, Zhihao Jia, et~al. 2025.
\newblock Magic{P}ig: {LSH} sampling for efficient llm generation.
\newblock \emph{Proceedings of ICLR}.

\bibitem[{Chu et~al.(2023)Chu, Chen, Chen, Yu, He, Wang, Peng, Liu, Qin, and Liu}]{chu2023survey}
Zheng Chu, Jingchang Chen, Qianglong Chen, Weijiang Yu, Tao He, Haotian Wang, Weihua Peng, Ming Liu, Bing Qin, and Ting Liu. 2023.
\newblock A survey of chain of thought reasoning: Advances, frontiers and future.
\newblock \emph{ArXiv:2309.15402}.

\bibitem[{Dao(2024)}]{dao2023flashattention2}
Tri Dao. 2024.
\newblock Flash{A}ttention-2: Faster attention with better parallelism and work partitioning.
\newblock \emph{Proceedings of ICLR}.

\bibitem[{Dao et~al.(2022)Dao, Fu, Ermon, Rudra, and R{\'e}}]{dao2022flashattention}
Tri Dao, Dan Fu, Stefano Ermon, Atri Rudra, and Christopher R{\'e}. 2022.
\newblock Flash{A}ttention: Fast and memory-efficient exact attention with io-awareness.
\newblock \emph{Proceedings of NeurIPS}.

\bibitem[{Dao and Gu(2024)}]{dao2024transformers}
Tri Dao and Albert Gu. 2024.
\newblock Transformers are {SSM}s: Generalized models and efficient algorithms through structured state space duality.
\newblock \emph{Proceedings of ICML}.

\bibitem[{DeepSeek-AI(2024)}]{deepseek-v3}
DeepSeek-AI. 2024.
\newblock Deepseek-{V}3 technical report.
\newblock \emph{ArXiv:2412.19437}.

\bibitem[{Dong et~al.(2024)Dong, Fu, Diao, Byeon, Chen, Mahabaleshwarkar, Liu, Van~Keirsbilck, Chen, Suhara et~al.}]{dong2024hymba}
Xin Dong, Yonggan Fu, Shizhe Diao, Wonmin Byeon, Zijia Chen, Ameya~Sunil Mahabaleshwarkar, Shih-Yang Liu, Matthijs Van~Keirsbilck, Min-Hung Chen, Yoshi Suhara, et~al. 2024.
\newblock Hymba: A hybrid-head architecture for small language models.
\newblock \emph{ArXiv:2411.13676}.

\bibitem[{Dubey et~al.(2024)Dubey, Jauhri, Pandey, Kadian, Al-Dahle, Letman, Mathur, Schelten, Yang, Fan et~al.}]{dubey2024llama3}
Abhimanyu Dubey, Abhinav Jauhri, Abhinav Pandey, Abhishek Kadian, Ahmad Al-Dahle, Aiesha Letman, Akhil Mathur, Alan Schelten, Amy Yang, Angela Fan, et~al. 2024.
\newblock The llama 3 herd of models.
\newblock \emph{ArXiv:2407.21783}.

\bibitem[{Ge et~al.(2024)Ge, Zhang, Liu, Zhang, Han, and Gao}]{ge2023model}
Suyu Ge, Yunan Zhang, Liyuan Liu, Minjia Zhang, Jiawei Han, and Jianfeng Gao. 2024.
\newblock Model tells you what to discard: Adaptive kv cache compression for llms.
\newblock \emph{Proceedings of ICLR}.

\bibitem[{Gu and Dao(2024)}]{gu2023mamba}
Albert Gu and Tri Dao. 2024.
\newblock Mamba: Linear-time sequence modeling with selective state spaces.
\newblock \emph{Proceedings of COLM}.

\bibitem[{He et~al.(2024)He, Zhang, Wu, Liu, Zhou, and Zhuang}]{he2024zipcache}
Yefei He, Luoming Zhang, Weijia Wu, Jing Liu, Hong Zhou, and Bohan Zhuang. 2024.
\newblock Zip{C}ache: Accurate and efficient kv cache quantization with salient token identification.
\newblock \emph{Proceedings of NeurIPS}.

\bibitem[{Hooper et~al.(2024)Hooper, Kim, Mohammadzadeh, Mahoney, Shao, Keutzer, and Gholami}]{hooper2024kvquant}
Coleman Hooper, Sehoon Kim, Hiva Mohammadzadeh, Michael~W Mahoney, Yakun~Sophia Shao, Kurt Keutzer, and Amir Gholami. 2024.
\newblock {KVQ}uant: Towards 10 million context length llm inference with kv cache quantization.
\newblock \emph{Proceedings of NeurIPS}.

\bibitem[{Hsieh et~al.(2024)Hsieh, Sun, Kriman, Acharya, Rekesh, Jia, Zhang, and Ginsburg}]{hsieh2024ruler}
Cheng-Ping Hsieh, Simeng Sun, Samuel Kriman, Shantanu Acharya, Dima Rekesh, Fei Jia, Yang Zhang, and Boris Ginsburg. 2024.
\newblock {RULER}: What's the real context size of your long-context language models?
\newblock \emph{Proceedings of COLM}.

\bibitem[{Huang et~al.(2019)Huang, Cheng, Bapna, Firat, Chen, Chen, Lee, Ngiam, Le, Wu, and Chen}]{huang2019gpipe}
Yanping Huang, Youlong Cheng, Ankur Bapna, Orhan Firat, Mia~Xu Chen, Dehao Chen, HyoukJoong Lee, Jiquan Ngiam, Quoc~V Le, Yonghui Wu, and Zhifeng Chen. 2019.
\newblock {GP}ipe: Efficient training of giant neural networks using pipeline parallelism.
\newblock \emph{Proceedings of NeurIPS}.

\bibitem[{Huang et~al.(2024)Huang, Yuan, Han, Xiao, and Liu}]{huang2024locret}
Yuxiang Huang, Binhang Yuan, Xu~Han, Chaojun Xiao, and Zhiyuan Liu. 2024.
\newblock Locret: Enhancing eviction in long-context llm inference with trained retaining heads.
\newblock \emph{ArXiv:2410.01805}.

\bibitem[{Jacobs et~al.(2023)Jacobs, Tanaka, Zhang, Zhang, Song, Rajbhandari, and He}]{jacobs2023deepspeed}
Sam~Ade Jacobs, Masahiro Tanaka, Chengming Zhang, Minjia Zhang, Shuaiwen~Leon Song, Samyam Rajbhandari, and Yuxiong He. 2023.
\newblock Deep{S}peed {U}lysses: System optimizations for enabling training of extreme long sequence transformer models.
\newblock \emph{ArXiv:2309.14509}.

\bibitem[{Jiang et~al.(2024)Jiang, Li, Zhang, Wu, Luo, Ahn, Han, Abdi, Li, Lin et~al.}]{jiang2024minference}
Huiqiang Jiang, Yucheng Li, Chengruidong Zhang, Qianhui Wu, Xufang Luo, Surin Ahn, Zhenhua Han, Amir~H Abdi, Dongsheng Li, Chin-Yew Lin, et~al. 2024.
\newblock M{I}nference 1.0: Accelerating pre-filling for long-context llms via dynamic sparse attention.
\newblock \emph{Proceedings of ICML}.

\bibitem[{Kim et~al.(2024)Kim, Kim, Choi, Park, Oh, and Park}]{kim2024survey}
Yeseung Kim, Dohyun Kim, Jieun Choi, Jisang Park, Nayoung Oh, and Daehyung Park. 2024.
\newblock A survey on integration of large language models with intelligent robots.
\newblock \emph{Intelligent Service Robotics}.

\bibitem[{Lee et~al.(2024)Lee, Lee, Seo, and Sim}]{lee2024infinigen}
Wonbeom Lee, Jungi Lee, Junghwan Seo, and Jaewoong Sim. 2024.
\newblock Infini{G}en: Efficient generative inference of large language models with dynamic {KV} cache management.
\newblock \emph{Proceedings of OSDI}.

\bibitem[{Li et~al.(2025)Li, Gong, Yang, Shan, Liu, Zhu, Zhang, Guo, Chen, Li et~al.}]{li2025minimax}
Aonian Li, Bangwei Gong, Bo~Yang, Boji Shan, Chang Liu, Cheng Zhu, Chunhao Zhang, Congchao Guo, Da~Chen, Dong Li, et~al. 2025.
\newblock Minimax-01: Scaling foundation models with lightning attention.
\newblock \emph{ArXiv preprint arXiv:2501.08313}.

\bibitem[{Li et~al.(2023)Li, Xue, Baranwal, Li, and You}]{li2021sequence}
Shenggui Li, Fuzhao Xue, Chaitanya Baranwal, Yongbin Li, and Yang You. 2023.
\newblock Sequence {P}arallelism: Long sequence training from system perspective.
\newblock \emph{Proceedings of ACL}.

\bibitem[{Li(2025)}]{li2024review}
Xinzhe Li. 2025.
\newblock A review of prominent paradigms for llm-based agents: Tool use (including {RAG}), planning, and feedback learning.
\newblock \emph{Proceedings of COLING}.

\bibitem[{Li et~al.(2024{\natexlab{a}})Li, Jiang, Wu, Luo, Ahn, Zhang, Abdi, Li, Gao, Yang et~al.}]{li2024scbench}
Yucheng Li, Huiqiang Jiang, Qianhui Wu, Xufang Luo, Surin Ahn, Chengruidong Zhang, Amir~H Abdi, Dongsheng Li, Jianfeng Gao, Yuqing Yang, et~al. 2024{\natexlab{a}}.
\newblock {SCB}ench: A kv cache-centric analysis of long-context methods.
\newblock \emph{ArXiv:2412.10319}.

\bibitem[{Li et~al.(2024{\natexlab{b}})Li, Huang, Yang, Venkitesh, Locatelli, Ye, Cai, Lewis, and Chen}]{li2024snapkv}
Yuhong Li, Yingbing Huang, Bowen Yang, Bharat Venkitesh, Acyr Locatelli, Hanchen Ye, Tianle Cai, Patrick Lewis, and Deming Chen. 2024{\natexlab{b}}.
\newblock Snap{KV}: Llm knows what you are looking for before generation.
\newblock \emph{Proceedings of NeurIPS}.

\bibitem[{Lieber et~al.(2025)Lieber, Lenz, Bata, Cohen, Osin, Dalmedigos, Safahi, Meirom, Belinkov, Shalev-Shwartz et~al.}]{lieber2024jamba}
Opher Lieber, Barak Lenz, Hofit Bata, Gal Cohen, Jhonathan Osin, Itay Dalmedigos, Erez Safahi, Shaked Meirom, Yonatan Belinkov, Shai Shalev-Shwartz, et~al. 2025.
\newblock Jamba: A hybrid transformer-mamba language model.
\newblock \emph{Proceedings of ICLR}.

\bibitem[{Liu et~al.(2024)Liu, Yuan, Jin, Zhong, Xu, Braverman, Chen, and Hu}]{liu2024kivi}
Zirui Liu, Jiayi Yuan, Hongye Jin, Shaochen Zhong, Zhaozhuo Xu, Vladimir Braverman, Beidi Chen, and Xia Hu. 2024.
\newblock {KIVI}: A tuning-free asymmetric 2bit quantization for {KV} cache.
\newblock \emph{Proceedings of ICML}.

\bibitem[{Lou et~al.(2024)Lou, Jia, Zheng, and Tu}]{lou2024sparser}
Chao Lou, Zixia Jia, Zilong Zheng, and Kewei Tu. 2024.
\newblock Sparser is {F}aster and {L}ess is {M}ore: Efficient sparse attention for long-range transformers.
\newblock \emph{ArXiv:2406.16747}.

\bibitem[{Narayanan et~al.(2021)Narayanan, Shoeybi, Casper, LeGresley, Patwary, Korthikanti, Vainbrand, Kashinkunti, Bernauer, Catanzaro et~al.}]{narayanan2021efficient}
Deepak Narayanan, Mohammad Shoeybi, Jared Casper, Patrick LeGresley, Mostofa Patwary, Vijay Korthikanti, Dmitri Vainbrand, Prethvi Kashinkunti, Julie Bernauer, Bryan Catanzaro, et~al. 2021.
\newblock Efficient large-scale language model training on gpu clusters using {M}egatron-{LM}.
\newblock \emph{Proceedings of SC}.

\bibitem[{OpenAI(2024)}]{gpt4o}
OpenAI. 2024.
\newblock \href {https://platform.openai.com/docs/models/gpt-4o} {Open{AI} {GPT}-4o}.

\bibitem[{Qin et~al.(2024)Qin, Hu, Lin, Chen, Ding, Cui, Zeng, Zhou, Huang, Xiao et~al.}]{qin2023tool}
Yujia Qin, Shengding Hu, Yankai Lin, Weize Chen, Ning Ding, Ganqu Cui, Zheni Zeng, Xuanhe Zhou, Yufei Huang, Chaojun Xiao, et~al. 2024.
\newblock Tool learning with foundation models.
\newblock \emph{ACM Computing Surveys}.

\bibitem[{Rajbhandari et~al.(2020)Rajbhandari, Rasley, Ruwase, and He}]{rajbhandari2020zero}
Samyam Rajbhandari, Jeff Rasley, Olatunji Ruwase, and Yuxiong He. 2020.
\newblock Ze{RO}: Memory optimizations toward training trillion parameter models.
\newblock \emph{Proceedings of SC}.

\bibitem[{Ren et~al.(2021)Ren, Rajbhandari, Aminabadi, Ruwase, Yang, Zhang, Li, and He}]{ren2021zero}
Jie Ren, Samyam Rajbhandari, Reza~Yazdani Aminabadi, Olatunji Ruwase, Shuangyan Yang, Minjia Zhang, Dong Li, and Yuxiong He. 2021.
\newblock Ze{RO}-{O}ffload: Democratizing billion-scale model training.
\newblock \emph{Proceedings of ATC}.

\bibitem[{Sahoo et~al.(2024)Sahoo, Singh, Saha, Jain, Mondal, and Chadha}]{sahoo2024systematic}
Pranab Sahoo, Ayush~Kumar Singh, Sriparna Saha, Vinija Jain, Samrat Mondal, and Aman Chadha. 2024.
\newblock A systematic survey of prompt engineering in large language models: Techniques and applications.
\newblock \emph{ArXiv:2402.07927}.

\bibitem[{Shah et~al.(2024)Shah, Bikshandi, Zhang, Thakkar, Ramani, and Dao}]{shah2024flashattention}
Jay Shah, Ganesh Bikshandi, Ying Zhang, Vijay Thakkar, Pradeep Ramani, and Tri Dao. 2024.
\newblock Flash{A}ttention-3: Fast and accurate attention with asynchrony and low-precision.
\newblock \emph{ArXiv:2407.08608}.

\bibitem[{Shi et~al.(2024)Shi, Zhang, Yao, Li, and Zhao}]{luohe2024keep}
Luohe Shi, Hongyi Zhang, Yao Yao, Zuchao Li, and Hai Zhao. 2024.
\newblock Keep the {C}ost {D}own: A review on methods to optimize llm's kv-cache consumption.
\newblock \emph{Proceedings of COLM}.

\bibitem[{Sun et~al.(2024{\natexlab{a}})Sun, Zhao, Han, Yang, Liu, Shi, and Sun}]{ao2024burstattention}
Ao~Sun, Weilin Zhao, Xu~Han, Cheng Yang, Zhiyuan Liu, Chuan Shi, and Maosong Sun. 2024{\natexlab{a}}.
\newblock Burst{A}ttention: An efficient distributed attention framework for extremely long sequences.
\newblock \emph{ArXiv:2403.09347}.

\bibitem[{Sun et~al.(2025)Sun, Zhao, Han, Yang, Zhang, Liu, Shi, and Sun}]{sun2024seq1f1b}
Ao~Sun, Weilin Zhao, Xu~Han, Cheng Yang, Xinrong Zhang, Zhiyuan Liu, Chuan Shi, and Maosong Sun. 2025.
\newblock Seq1{F}1{B}: Efficient sequence-level pipeline parallelism for large language model training.
\newblock \emph{Proceedings of NAACL}.

\bibitem[{Sun et~al.(2024{\natexlab{b}})Sun, Chang, Bao, Zheng, Zheng, Liu, Dong, Chi, and Chen}]{sun2024shadowkv}
Hanshi Sun, Li-Wen Chang, Wenlei Bao, Size Zheng, Ningxin Zheng, Xin Liu, Harry Dong, Yuejie Chi, and Beidi Chen. 2024{\natexlab{b}}.
\newblock Shadow{KV}: {KV} cache in shadows for high-throughput long-context llm inference.
\newblock \emph{ArXiv:2410.21465}.

\bibitem[{Vaswani(2017)}]{vaswani2017attention}
Ashish Vaswani. 2017.
\newblock Attention is all you need.
\newblock \emph{Proceedings of NeurIPS}.

\bibitem[{Wan et~al.(2024)Wan, Wang, Liu, Alam, Zheng et~al.}]{wan2023efficient}
Zhongwei Wan, Xin Wang, Che Liu, Samiul Alam, Yu~Zheng, et~al. 2024.
\newblock Efficient large language models: A survey.
\newblock \emph{Transactions on Machine Learning Research}.

\bibitem[{Wang et~al.(2024)Wang, Jin, Yu, and Zhang}]{wang2024model}
Zheng Wang, Boxiao Jin, Zhongzhi Yu, and Minjia Zhang. 2024.
\newblock Model tells you where to merge: Adaptive kv cache merging for llms on long-context tasks.
\newblock \emph{arXiv:2407.08454}.

\bibitem[{Xiao et~al.(2024{\natexlab{a}})Xiao, Zhang, Han, Xiao, Lin, Zhang, Liu, Han, and Sun}]{xiao2024infllm}
Chaojun Xiao, Pengle Zhang, Xu~Han, Guangxuan Xiao, Yankai Lin, Zhengyan Zhang, Zhiyuan Liu, Song Han, and Maosong Sun. 2024{\natexlab{a}}.
\newblock Inf{LLM}: Unveiling the intrinsic capacity of llms for understanding extremely long sequences with training-free memory.
\newblock \emph{Proceedings of NeurIPS}.

\bibitem[{Xiao et~al.(2024{\natexlab{b}})Xiao, Tian, Chen, Han, and Lewis}]{xiao2023efficient}
Guangxuan Xiao, Yuandong Tian, Beidi Chen, Song Han, and Mike Lewis. 2024{\natexlab{b}}.
\newblock Efficient streaming language models with attention sinks.
\newblock \emph{Proceedings of ICLR}.

\bibitem[{Yang et~al.(2025)Yang, Chen, and Chen}]{yang2025ape}
Xinyu Yang, Tianqi Chen, and Beidi Chen. 2025.
\newblock {APE}: Faster and longer context-augmented generation via adaptive parallel encoding.
\newblock \emph{Proceedings of ICLR}.

\bibitem[{Yao et~al.(2024)Yao, Li, and Zhao}]{yao2024sirllm}
Yao Yao, Zuchao Li, and Hai Zhao. 2024.
\newblock Sir{LLM}: Streaming infinite retentive {LLM}.
\newblock \emph{Proceedings of ACL}.

\bibitem[{Zaheer et~al.(2020)Zaheer, Guruganesh, Dubey, Ainslie, Alberti, Ontanon, Pham, Ravula, Wang, Yang, and Ahmed}]{zaheer2020big}
Manzil Zaheer, Guru Guruganesh, Kumar~Avinava Dubey, Joshua Ainslie, Chris Alberti, Santiago Ontanon, Philip Pham, Anirudh Ravula, Qifan Wang, Li~Yang, and Amr Ahmed. 2020.
\newblock Big {B}ird: Transformers for longer sequences.
\newblock \emph{Proceedings of NeurIPS}.

\bibitem[{Zeng et~al.(2023)Zeng, Gan, Wang, Liu, and Yu}]{zeng2023large}
Fanlong Zeng, Wensheng Gan, Yongheng Wang, Ning Liu, and Philip~S Yu. 2023.
\newblock Large language models for robotics: A survey.
\newblock \emph{ArXiv:2311.07226}.

\bibitem[{Zhang et~al.(2024{\natexlab{a}})Zhang, Yi, Xu, and Shrivastava}]{zhang2024kv}
Tianyi Zhang, Jonah Yi, Zhaozhuo Xu, and Anshumali Shrivastava. 2024{\natexlab{a}}.
\newblock {KV} cache is 1 bit per channel: Efficient large language model inference with coupled quantization.
\newblock \emph{Proceedings of NeurIPS}.

\bibitem[{Zhang et~al.(2024{\natexlab{b}})Zhang, Chen, Hu, Xu, Chen, Hao, Han, Thai, Wang, Liu et~al.}]{zhang2024infty}
Xinrong Zhang, Yingfa Chen, Shengding Hu, Zihang Xu, Junhao Chen, Moo~Khai Hao, Xu~Han, Zhen~Leng Thai, Shuo Wang, Zhiyuan Liu, et~al. 2024{\natexlab{b}}.
\newblock $\infty $ bench: Extending long context evaluation beyond 100k tokens.
\newblock \emph{Proceedings of ACL}.

\bibitem[{Zhang et~al.(2024{\natexlab{c}})Zhang, Du, Du, Pang, Gao, and Lin}]{zhang2024simlayerkv}
Xuan Zhang, Cunxiao Du, Chao Du, Tianyu Pang, Wei Gao, and Min Lin. 2024{\natexlab{c}}.
\newblock Sim{L}ayer{KV}: A simple framework for layer-level kv cache reduction.
\newblock \emph{ArXiv:2410.13846}.

\bibitem[{Zhang et~al.(2024{\natexlab{d}})Zhang, Du, Luo, Zhong, Zhang, Liu, and Ji}]{zhangcam}
Yuxin Zhang, Yuxuan Du, Gen Luo, Yunshan Zhong, Zhenyu Zhang, Shiwei Liu, and Rongrong Ji. 2024{\natexlab{d}}.
\newblock Ca{M}: Cache merging for memory-efficient llms inference.
\newblock In \emph{Proceedings of ICML}.

\bibitem[{Zhang et~al.(2024{\natexlab{e}})Zhang, Sheng, Zhou, Chen, Zheng, Cai, Song, Tian, R{\'e}, Barrett et~al.}]{zhang2024h2o}
Zhenyu Zhang, Ying Sheng, Tianyi Zhou, Tianlong Chen, Lianmin Zheng, Ruisi Cai, Zhao Song, Yuandong Tian, Christopher R{\'e}, Clark Barrett, et~al. 2024{\natexlab{e}}.
\newblock H$_2${O}: Heavy-hitter oracle for efficient generative inference of large language models.
\newblock \emph{Proceedings of NeurIPS}.

\bibitem[{Zhao et~al.(2024)Zhao, Zu, Xu, Lu, He, Ding, Gui, Zhang, and Huang}]{zhao2024longagent}
Jun Zhao, Can Zu, Hao Xu, Yi~Lu, Wei He, Yiwen Ding, Tao Gui, Qi~Zhang, and Xuanjing Huang. 2024.
\newblock Long{A}gent: Scaling language models to 128k context through multi-agent collaboration.
\newblock \emph{Proceedings of EMNLP}.

\bibitem[{Zhao et~al.(2023)Zhao, Zhou, Li, Tang, Wang, Hou, Min, Zhang, Zhang, Dong et~al.}]{zhao2023survey}
Wayne~Xin Zhao, Kun Zhou, Junyi Li, Tianyi Tang, Xiaolei Wang, Yupeng Hou, Yingqian Min, Beichen Zhang, Junjie Zhang, Zican Dong, et~al. 2023.
\newblock A survey of large language models.
\newblock \emph{ArXiv:2303.18223}.

\end{thebibliography}

\clearpage

\appendix

\section{Pseudocode of \name}

\begin{algorithm}[h]
    \small
    \DontPrintSemicolon
    \caption{\name~Inference}
    \label{alg:infer}
    \KwIn{Model $\mathbf{M}$, Input sequence $t$, Anchor length $l_a$, Passing length $l_p$, Host index $h$, Total host number $H$, Stop Criteria \texttt{SC}}
    \KwOut{Generated tokens $t_{gen}$}

    $d$, $q$ $\leftarrow$ \texttt{SplitDocumentQuery}($t$)\;\label{line:split}
    $\cblock{\mathbf{B}}_h$ $\leftarrow$ \texttt{SplitContext}($d$, $h$)\;\label{line:index}
    \tcp{Make anchor block and embed query}
    $\canchor{\mathbf{A}}$ $\leftarrow$ $[\canchor{q_1, \cdots, q_{l_q},~d_{1}, \cdots, d_{l_a}}]$\;\label{line:setbeg}
    \If{$h=1$}{$\canchor{\mathbf{A}}\leftarrow \texttt{None}$}\label{line:setend}
    $\canchor{\mathbf{H}}^{(0)}_a, \cblock{\mathbf{H}}_h^{(0)}$ $\leftarrow$ $\mathbf{M}$\texttt{.embedding}($\mathbf{A}, \mathbf{B}_h$)\;\label{line:embed}
    \color{blue}
    \tcp{{Prefill}}
    \color{black}
    \texttt{KV\_cache} $\leftarrow [~]$\;\label{line:prefillbeg}
    \For {$i$,~\texttt{l} $\in \texttt{enumerate(}\mathbf{M}$\texttt{.layers)}}{
        $\canchor{\mathbf{H}}^{(i)}_a, \cblock{\mathbf{H}}_h^{(i)}, \canchor{\mathbf{K}}^{(i)}_a, \cblock{\mathbf{K}}_h^{(i)},\canchor{\mathbf{V}}^{(i)}_a, \cblock{\mathbf{V}}_h^{(i)}$ $\leftarrow$ \hyperref[alg:apb_prefill]{\texttt{\underline{APB}}}($\texttt{l}, ~\canchor{\mathbf{H}}^{(i-1)}_a, ~\cblock{\mathbf{H}}_h^{(i-1)}, ~l_p, ~h, ~H$)\;
        \texttt{KV\_cache.append}($\cblock{\mathbf{K}}_h^{(i)},~\cblock{\mathbf{V}}_h^{(i)}$)
    }\label{line:prefillend}
    
    \color{blue}
    \tcp{{Decoding}}
    \color{black}
    $t_{gen}\leftarrow [~]$\;\label{line:decodebeg}
    $x\leftarrow q$\;
    \While {\rm{not} \texttt{SC}($\mathbf{M}, t, ~t_{gen}$)}{
        $\mathbf{H}^{(0)}\leftarrow \mathbf{M}$\texttt{.embedding}($x$)\;
        \For {$i$,~\texttt{l} $\in \texttt{enumerate(}\mathbf{M}$\texttt{.layers)}}{
            $\mathbf{H}^{(i)}, \mathbf{K}^{(i)}, \mathbf{V}^{(i)}$ $\leftarrow$ \hyperref[alg:apb_decode]{\texttt{\underline{Accu}}}($\texttt{l}, ~\mathbf{H}^{(i-1)}, ~\texttt{KV\_cache}[i],~h, ~H$)\;
            \If{$h=H$}{
                \texttt{KV\_cache}[$i$]\texttt{.extend}($\mathbf{K}^{(i)},~\mathbf{V}^{(i)}$)\;
            }
        }
        $x\leftarrow$ \texttt{GenerateNextToken}($\mathbf{H}^{(L)}, ~\mathbf{M}\texttt{.LM\_Head}$)\;
        $t_{gen}\leftarrow[t_{gen}, x]$\;
    }\label{line:decodeend}
    \Return $t_{gen}$\;\label{line:ret}
\end{algorithm}

\begin{algorithm}[h]
    \small
    \DontPrintSemicolon
    \caption{\name~Prefill Function}
    \label{alg:apb_prefill}
    \KwIn{Layer \texttt{layer}, Anchor hidden states $\canchor{\mathbf{H}}_a$, Block hidden states $\cblock{\mathbf{H}}_h$, Passing length $l_p$, Host index $h$, Total host number $H$}
    \KwOut{Output hidden states $\canchor{\mathbf{H}}_a^{out}, \cblock{\mathbf{H}}_h^{out}$, KV cache $\canchor{\mathbf{K}}_a, \cblock{\mathbf{K}}_h,\canchor{\mathbf{V}}_a, \cblock{\mathbf{V}}_h$}

    $\left[\canchor{\mathbf{Q}}_a, \cblock{\mathbf{Q}}_h\right], \left[\canchor{\mathbf{K}}_a, \cblock{\mathbf{K}}_h\right], \left[\canchor{\mathbf{V}}_a, \cblock{\mathbf{V}}_h\right]\leftarrow \texttt{layer.qkv\_proj}(\left[\canchor{\mathbf{H}}_a, \cblock{\mathbf{H}}_h\right])$ \;
    \color{blue}
    \tcp{Compress KV cache by Retaining Head}
    \color{black}
    $s_1, \cdots, s_{l_b} \leftarrow \texttt{layer.}\mathcal{R}(\left[\cblock{\mathbf{Q}}_h, \cblock{\mathbf{K}}_h, \cblock{\mathbf{V}}_h\right])$\;\label{line:retbeg}
    \texttt{indices} $\leftarrow$ ArgTop-$l_p$($s_1, \cdots, s_{l_b}$)\;
    $\cpass{\mathbf{K}}_h^C, \cpass{\mathbf{V}}_h^C\leftarrow \cblock{\mathbf{K}}_h[\texttt{indices}], \cblock{\mathbf{V}}_h[\texttt{indices}]$\;\label{line:retend}

    \color{blue}
    \tcp{Communication}
    \color{black}
    $\cpass{\mathbf{K}}_1^C, \cdots, \cpass{\mathbf{K}}_H^C, \leftarrow \texttt{AllGather}(\cpass{\mathbf{K}}_h^C)$\;\label{line:commbeg}
    $\cpass{\mathbf{V}}_1^C, \cdots, \cpass{\mathbf{V}}_H^C, \leftarrow \texttt{AllGather}(\cpass{\mathbf{V}}_h^C)$\;\label{line:commend}

    $\cpass{\mathbf{K}}_p \leftarrow \cpass{\mathbf{K}}_1^C, \cdots, \cpass{\mathbf{K}}_{h-1}^C$
    $\cpass{\mathbf{V}}_p \leftarrow \cpass{\mathbf{V}}_1^C, \cdots, \cpass{\mathbf{V}}_{h-1}^C$\;\label{line:pass}

    \color{blue}
    \tcp{Attention with $\mathbf{A}$ block and $\mathbf{P}$ block}
    \color{black}
    $\canchor{\mathbf{A}}_a, \cblock{\mathbf{A}}_h\leftarrow \texttt{Attention}([\canchor{\mathbf{Q}}_a, \cblock{\mathbf{Q}}_h], [\canchor{\mathbf{K}}_a, \cpass{\mathbf{K}}_p, \cblock{\mathbf{K}}_h], [\canchor{\mathbf{V}}_a, \cpass{\mathbf{V}}_p, \cblock{\mathbf{V}}_h])$\label{line:attn}

    $\canchor{\mathbf{H}}_a^{out}, \cblock{\mathbf{H}}_h^{out}\leftarrow \texttt{FFN}(\canchor{\mathbf{A}}_a, \cblock{\mathbf{A}}_h)$\;\label{line:ffn}
    
    \Return $\canchor{\mathbf{H}}_a^{out}, ~\cblock{\mathbf{H}}_h^{out}, ~\canchor{\mathbf{K}}_a, ~\cblock{\mathbf{K}}_h, ~\canchor{\mathbf{V}}_a, ~\cblock{\mathbf{V}}_h$\;
\end{algorithm}

\begin{algorithm}[h]
    \small
    \DontPrintSemicolon
    \caption{\name~Decoding Function}
    \label{alg:apb_decode}
    \KwIn{Layer \texttt{layer}, Hidden states $\mathbf{H}$, KV cache \texttt{cache}, Host index $h$, Total host number $H$}
    \KwOut{Output hidden states $\mathbf{H}^{out}$, KV cache $\mathbf{K},~\mathbf{V}$}

    $\mathbf{Q},~\mathbf{K},~\mathbf{V}\leftarrow \texttt{layer.qkv\_proj}(\mathbf{H})$ \;\label{line:qkv}
    $\mathbf{K}_{cache},~\mathbf{V}_{cache}\leftarrow \texttt{cache}$\;
    \If{$h<H$}{
        $\mathbf{A}, \texttt{lse}\leftarrow \texttt{Attention}(\mathbf{Q}, ~\mathbf{K}_{cache}, ~\mathbf{V}_{cache})$\;\label{line:attnl}
    }\Else{
        $\mathbf{A}, \texttt{lse}\leftarrow \texttt{Attention}(\mathbf{Q}, ~[\mathbf{K}_{cache}, \mathbf{K}], ~[\mathbf{V}_{cache}, \mathbf{V}])$\;\label{line:attnh}
    }

    $\mathbf{A}_1, \cdots, \mathbf{A}_H;\texttt{lse}_1, \cdots, \texttt{lse}_H\leftarrow \texttt{Gather}(\mathbf{A}, \texttt{lse})$\;\label{line:gather}

    $\mathbf{A}\leftarrow\texttt{MergeScore}(\mathbf{A}_1, \cdots, \mathbf{A}_H;\texttt{lse}_1, \cdots, \texttt{lse}_H)$\label{line:merge}

    $\mathbf{H}^{out}\leftarrow \texttt{FFN}(\mathbf{A})$\label{line:ffn2}
    
    \Return $\mathbf{H}^{out}, ~\mathbf{K}, ~\mathbf{V}$\;
\end{algorithm}

Here, we present the complete inference process of \name~in Algorithm~\ref{alg:infer}.

For initialization, the input sequence $t$ is first split into the document $d$ and query $q$ (line \ref{line:split}). 
Each host then takes a partial context $\cblock{\textbf{B}}_h$ based on the host index $h$ (line \ref{line:index}) and sets the anchor block (lines \ref{line:setbeg}--\ref{line:setend}). 
To begin prefill, the embeddings of the anchor block $\canchor{\textbf{A}}$ and the context block $\cblock{\textbf{B}}_h$ are computed (line \ref{line:embed}). 
During the prefill stage (lines \ref{line:prefillbeg}--\ref{line:prefillend}), we employ Algorithm~\ref{alg:apb_prefill} to perform sequence parallelism-aware approximate attention. 
In the decoding stage (lines \ref{line:decodebeg}--\ref{line:decodeend}), we directly apply accurate attention with online softmax (Algorithm~\ref{alg:apb_decode}), as introduced in \textsc{StarAttn}~\citep{acharya2024star} as \textit{stage-2}, to generate new tokens. 
Finally, the generated tokens are returned (line \ref{line:ret}).

The prefill stage function of \name~is described in Algorithm~\ref{alg:apb_prefill}. 
Before performing the attention calculation, the KV cache of the current host is compressed using the retaining heads $\mathcal{R}$ (line \ref{line:retbeg}--\ref{line:retend}).  
Next, two \texttt{AllGather} communications are performed on the compressed K and V caches, allowing each host to obtain a full view of the compressed KV cache from all hosts (line \ref{line:commbeg}--\ref{line:commend}).  
Subsequently, the passing block is formed by concatenating the compressed KV caches sent from the previous hosts (line \ref{line:pass}), followed by the modified attention mechanism (line \ref{line:attn}).  
After the attention calculation, the passing blocks are discarded, and only the attention scores corresponding to the anchor block and the context block held by the current host are passed through the \texttt{FFN} (line \ref{line:ffn}).

The decoding function of \name, also known as \textit{stage-2} of \textsc{StarAttn}, is described in Algorithm~\ref{alg:apb_decode}.  
In the decoding stage, all hosts hold the hidden states $\textbf{H}$ of the new token.  
The $\textbf{Q}$, $\textbf{K}$, and $\textbf{V}$ of the new token are computed simultaneously on each host (line \ref{line:qkv}).  
For all hosts except the last one, the attention score is calculated by the new token's query with the KV cache held by the current host (line \ref{line:attnl}).  
For the last host, the local KV cache is first concatenated with the new KV, and then attention calculation is performed with the query (line \ref{line:attnh}).  
A \texttt{Gather} communication is performed to collect  $\textbf{A}_h$ and \texttt{lse} calculated by each host (line \ref{line:gather}), which are then merged to obtain the global attention score $\textbf{A}$ (line \ref{line:merge}).  
Notably, the $\textbf{A}$ on each host should be identical.  
Finally, $\textbf{A}$ is passed through the \texttt{FFN} simultaneously on each host, generating identical $\textbf{H}^{out}$ on all hosts (line \ref{line:ffn2}).

\section{Reproducibility Settings}
\label{appendix:hyper}

\subsection{Training of the Compressor}

We use the retaining heads proposed in~\citet{huang2024locret} as our compressor.
Following~\citet{huang2024locret}, we set the intermediate size of the retaining heads to 1024.
We utilize the first 3000 samples from the LongAlign~\citep{bai2024longalign} dataset to generate the training labels, and we train the retaining heads with the a frozen model backbone for 3000 steps in 1 epoch.
The batch size set is 1 and the maximum input length is set to 10240. We use AdamW as the optimizer and the learning rate is set to 5e-4, with  $\beta_1$ set to 0.9 and $\beta_2$ set to 0.95. 
We apply a linear scheduler whose number of warmup steps is set to 300.
Notably, the training loss of the retaining head consists of a regression loss and a smoothing loss, following the setting of~\cite {huang2024locret}, and the balance factor $\alpha$ is set to 0.0025.
The gradient clipping value is set to 0.5 to avoid gradient explosion.

\begin{table}[h]
\small
\center
\begin{tabular}{l|ccccc}
\toprule
$n$ & 32K & 64K & 128K & 256K & 512K \\
\midrule
$l_b$   & 4K  & 8K  & 16K  & 32K  & 64K  \\
$l_a$ & 1K  & 2K  & 4K   & 8K   & 8K\\
$l_p$ & 0.5K& 1K& 2K& 4K& 8K\\
\bottomrule
\end{tabular}
\caption{Hyperparameters of \name~used in Section~\ref{sec:varlen}. $n$ stands for input length, $l_b$ is the block size, $l_a$ represents the anchor length, and $l_p$ represents the passing length. ``K'' is an abbreviation for 1024.}

\label{tab:seq_params}
\end{table}

\subsection{Inference Hyperparameters}

Here, we elaborate on the details of the inference hyperparameters used in Section~\ref{sec:end-to-end}, ~\ref{sec:varlen}, and~\ref{sec:abl}.

\subsubsection{End-to-End Benchmark (Section~\ref{sec:end-to-end})}

In the end-to-end benchmark, we first test the performance of 3 baselines with the performance of our method. 
Since methods without approximate attention do not alter the computational outcome of the attention mechanism, we take the result of \textsc{Ulysses} as the performance benchmark for \textsc{FullAttn}.
Then, we evaluate the inference speed of all baselines along with \name~.

\textbf{Performance Evaluation.}
In the $\infty$Bench experiments, we evaluate the performance of each method by running all the data from each task. 
Details of the tasks can be found in \citet{zhang2024infty}. 
When conducting the RULER experiments, we generate 500 test samples for each RULER task to evaluate the performance, with details introduced in \citet{hsieh2024ruler}.
We report the performance of each method under a 128K input length for both benchmarks. 
For methods with sequence parallelism, we set the sequence parallel size to 8.
We use a single machine with 8 GPUs for \texttt{Llama-3.1-8B} and \texttt{Qwen-2.5-14B}, while for \texttt{Yi-34B}, due to its large model size, we employ two machines with layers evenly distributed between them.
All the answers are generated with greedy decoding, i.e. with a temperature set to 0.
For \textsc{StarAttn}, we set both the block length and the anchor length to 16K.
For \textsc{MInference}, we use the official head configuration for \texttt{Llama-3.1-8B} and generate the configurations for \texttt{Qwen-2.5-14B} and \texttt{Yi-34B} based on the first data entry of SG1 from RULER.
For \name, we set anchor length $l_a$ to 4K and passing length $l_p$ to 2K. 
We place the query tokens after the system prompt to embed the query within the anchor blocks without disrupting the model's chat template.

\textbf{Speed Evaluation.}
We run the first 20 samples for each task and average the inference speed. 
To reflect the speed of LLMs in processing a query, we define the inference speed as follows.
\begin{align*}
    \texttt{speed} = \frac{\#\texttt{input tokens} + \#\texttt{output tokens}}{\texttt{prefill time} + \texttt{decoding time}}
\label{eq:speed}
\end{align*}
We keep all the hyperparameters and distribution settings same as the performance evaluation.

\subsubsection{Benchmarking Across Input Lengths (Section~\ref{sec:varlen})}

For Section 4.3, we evaluate the performance, speed, and report the compute of all the methods across different input lengths using the \texttt{Llama-3.1-8B} model and the RULER benchmark. 
We run 50 samples per task for performance evaluation, while for speed testing, we run 20 samples per task. 
For \textsc{StarAttn}, we set both the block size and the anchor length to $n/H$, where $n$ is the input length and $H$ is the sequence parallelism size.
For \name, we use the hyperparameters listed in Table~\ref{tab:seq_params}.
We align other hyperparameters to the end-to-end benchmark.
We visualize the compute of each method following Table~\ref{tab:flops}.

\subsubsection{Ablation Studies (Section~\ref{sec:abl})}

Here, we elaborate on the settings in Section~\ref{sec:abl}.

\textbf{Each Component's Contribution.}
We conduct the ablation study of each component using \texttt{Llama-3.1-8B}.
We conduct experiments of different settings on E.MC from $\infty$Bench using the first 50 samples to assess the impact of different combinations on performance. 
We use the hyperparameters of $n=128\text{K},~l_b=32\text{K},~l_a=4\text{K},~l_p=2\text{K}$.

\textbf{Wall-time Breakdown Analysis.}
The prefill and decoding time is tested with \texttt{Llama-3.1-8B} on SG1 from RULER. 
For the breakdown analysis of the prefill time, we run all the methods on synthetic random input.
All the experiments are conducted under a 128K input length, and the hyperparameters are kept the same as in previous experiments.

\textbf{Distributed Settings.}
We test the performance under different host numbers (sequence parallelism size) with \texttt{Llama-3.1-8B}. 
We set $H$ across $\{2,4,6,8\}$ and evaluate the performance of \textsc{StarAttn} and \name~on E.MC from $\infty$Bench.
We set the sequence length to 32K and 128K, and test on the first 50 samples.

\subsection{System Environment}

All training and evaluation are conducted on workstations equipped with 8$\times$ \texttt{NVIDIA A800-80GB} GPUs and \texttt{104 Intel® Xeon® Platinum 8470} CPUs, running \texttt{CentOS Linux 7 (Core)}.
The GPUs within each machine are interconnected using third-generation NVLink technology, while cross-machine communication is facilitated via HDR InfiniBand.
We perform the training of the retaining head and all experiments—except for \textsc{FlashAttn} and \textsc{MInference} tests—on 8 GPUs. The experiments for \textsc{FlashAttn} and \textsc{MInference} are conducted on a single GPU.

\section{Baselines}
\label{appendix:baselines}

In this section, we briefly introduce the design principles of each baseline.

\textbf{\textsc{FlashAttn}.} \citet{dao2023flashattention2} introduces an attention computation method with hardware awareness that leverages the architecture of modern GPUs. \textsc{FlashAttn} applies tiling techniques to the calculation of attention scores and fully utilizes the high-bandwidth memory to accelerate computation. \textsc{FlashAttn} is an accurate attention method that preserves the computation results, primarily focusing on accelerating attention on a single GPU.

\textbf{\textsc{MInference}.} 
As introduced in \citet{jiang2024minference}, three approximate attention patterns can be utilized to approximate \textsc{FullAttn} computation. \textsc{MInference} first searches for the head configuration by assigning an approximate pattern to each head. During inference, only a limited number of attention score entries are calculated, accelerating the prefill by reducing the computation in attention. \textsc{MInference} focuses on applying approximate attention without sequence parallelism.

\textbf{\textsc{RingAttn}.} 
Introduced by \citet{li2021sequence}, \textsc{RingAttn} distributes the context across multiple GPUs (hosts). Online softmax is applied to the attention calculation, where each host computes the partial attention score passed by the previous host, using local context for $H-1$ rounds ($H$ is the number of hosts). The core idea is to overlap the communication of passing partial attention scores with the attention calculation. \textsc{RingAttn} is primarily used for training LLMs with extremely long contexts but is not optimized for inference.

\textbf{\textsc{Ulysses}.} 
Similar to \textsc{RingAttn}, \textsc{Ulysses}~\citep{jacobs2023deepspeed} was also introduced for long-context training. \textsc{Ulysses} applies sequence parallelism, with each host holding a partial context. During attention calculation, three \texttt{AlltoAll} communications on $\textbf{Q}$, $\textbf{K}$, and $\textbf{V}$ are performed to distribute the full context of specific heads to the corresponding host. After the calculation, another \texttt{AlltoAll} communication is conducted on the attention score to revert the distribution from splitting heads to splitting the sequence.

\textbf{\textsc{StarAttn}.}
Introduced by \citet{acharya2024star}, \textsc{StarAttn} is a pioneer in combining approximate attention with sequence parallelism. \textsc{StarAttn} introduces \textit{anchor blocks}, which contain the initial tokens of the input sequence and are the same size as the blocks on each host. Each host calculates only the partial attention between the anchor block and the current context block, with no communication required. Although \textsc{StarAttn} reduces communication and synchronization, its large anchor blocks introduce overhead in the \texttt{FFN}, limiting the speedup gains.

\begin{table*}[t]
\small
\center
\scalebox{0.87}{
\begin{tabular}{l|c}
\toprule
Method & FLOPs / forward \\
\midrule
\textsc{FullAttn} & $L\times\left(4nd^2+\frac{4}{g}nd^2+2n^2d+6ndI\right)$ \\
\textsc{StarAttn} & $\frac{L}{H}\times\left[(8H-4)nd^2+\frac{8H-4}{g}nd^2+\frac{8H-6}{H}n^2d+(12H-6)ndI\right]$ \\
\name & $L\times \left[4 \left(1 + \frac{1}{g} + \frac{0.5n}{Hd} + \frac{1.5I}{d}\right) \frac{n}{H}d^2 + 4  (H-1) \left(1 + \frac{1}{g} + \frac{0.5\left(\frac{n}{H} + l_a\right)}{d} + \frac{1.5I}{d}\right)\left(\frac{n}{H} + l_a\right)d^2 +   l_p H (H-1)\left(\frac{n}{H} + l_a\right)d\right]$ \\
\bottomrule
\end{tabular}}
\vspace{-3pt}
\caption{The FLOPs per forward call of \textsc{FullAttn}, \textsc{StarAttn} and \name. ``$L$'' stands for the number of layers, ``$n$'' stands for the input length, ``$d$'' is the hidden size of the model, and ``$I$'' is the intermediate size of \texttt{FFN}. ``$H$'' stands for the number of hosts, ``$l_a$'' is the anchor length, and ``$l_p$'' is the passing length. We calculate the compute of the model without the input embedding, the language-modeling head, positional embeddings, and all the normalizations.}
\vspace{-5pt}

\label{tab:flops}
\end{table*}

Here, we present the computation per forward call for each method in Table~\ref{tab:flops}. Notably, since \textsc{FlashAttn}, \textsc{RingAttn}, and \textsc{Ulysses} are \textsc{FullAttn} methods, they share the same computation formula. We do not include \textsc{MInference}, as its computation depends on the search results of the head configurations.
We provide a visualization of Table~\ref{tab:flops} in Figure~\mbox{\ref{fig:var-seqlen}\hspace{-20pt}\subref{fig:var-seqlen-c}}.

\section{Task Abbreviations}
\label{sec:abbre}

Here, we provide a mapping between the benchmark task abbreviations and their full names used in the main text in Table~\ref{tab:name_map_inf} and~\ref{tab:name_map_ruler}.

\begin{table}[H]
\small
\center
\scalebox{0.9}{
\begin{tabular}{c|c}
\toprule
Abbr. & Full Name \\
\midrule
R.PassKey & Retrieve.PassKey \\
R.Number & Retrieve.Number \\
R.KV & Retrieve.KV \\
E.Sum & En.Sum \\
E.QA & En.QA \\
E.Dia & En.Dia \\
Z.QA & Zh.QA \\
C.Debug & Code.Debug \\
M.Find & Math.Find \\
\bottomrule
\end{tabular}}
\caption{Task name mapping of $\infty$Bench.}
\vspace{-10pt}

\label{tab:name_map_inf}
\end{table}

\begin{table}[H]
\small
\center
\scalebox{0.9}{
\begin{tabular}{c|c}
\toprule
Abbr. & Full Name \\
\midrule
SG1 & Single NIAH 1 \\
SG2 & Single NIAH 2 \\
SG3 & Single NIAH 3 \\
MK1 & Multi-keys NIAH 1 \\
MK2 & Multi-keys NIAH 2 \\
MK3 & Multi-keys NIAH 3 \\
MV & Multi-values NIAH \\
MQ & Multi-queries NIAH \\
VT & Variable Tracking \\
CWE & Common Words Extraction \\
FWE & Frequent Words Extraction \\
QA1 & Question Answering 1 \\
QA2 & Question Answering 2 \\
\bottomrule
\end{tabular}}
\caption{Task name mapping of RULER.}

\label{tab:name_map_ruler}
\end{table}

\section{Further Ablation Studies and Detailed Experimental Results}
\label{sec:detailed_results}
Here, we represent the further ablation studies in Section~\ref{sec:break_pd} and ~\ref{sec:hyper_stability}. We list all experimental data that are not included in the main text in Section~\ref{sec:detail-results}. 

\subsection{Breakdown Analysis: Prefill and Decoding}
\label{sec:break_pd}
\begin{figure}[H]
\begin{center}
\includegraphics[width=0.8\linewidth]{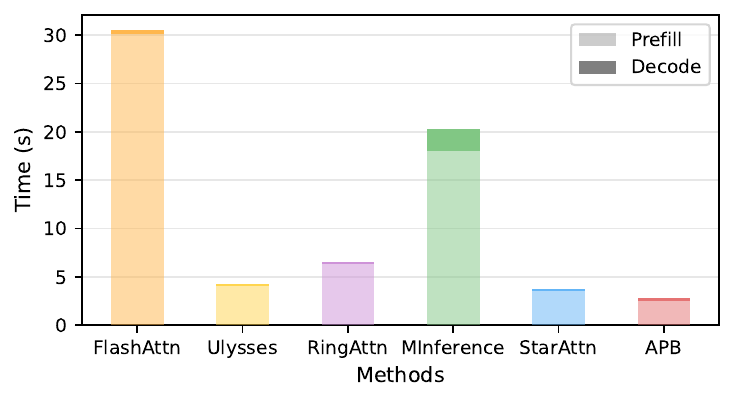}
\end{center}
\caption{The wall-time of prefill and decoding for various methods on 128K context. Prefill is the major bottleneck in long-context query processing.}
\label{fig:breakdown-pd}
\end{figure}

We conduct a wall-time breakdown analysis to further investigate how \name~achieves a higher inference speed. First, we measure the prefill and decoding times for each method, and the results shown in Figure~\ref{fig:breakdown-pd} reveal that the prefill stage is the bottleneck, while the decoding stage takes up significantly less time. Since \name~optimizes the prefill stage, it is able to address the most time-consuming part of the process. 

\subsection{Hyperparameter Stability}
\label{sec:hyper_stability}
There are two hyperparameters introduced in the \name~framework: the length of the anchor block $l_a$, and the passing length $l_p$. In this study, we examine the sensitivity of these two parameters by measuring the performance on the E.QA task from $\infty$Bench for various lengths of $l_a$ and $l_p$.

In this experiment, we use \texttt{Llama-3.1-8B} as the model backbone.
We test on the first 50 samples and set the sequence length to 128K.
We present the results with $l_a, l_p\in \{1\text{K},2\text{K},3\text{K},4\text{K}\}$.

\begin{figure}[h]
\begin{center}
\includegraphics[width=0.9\linewidth]{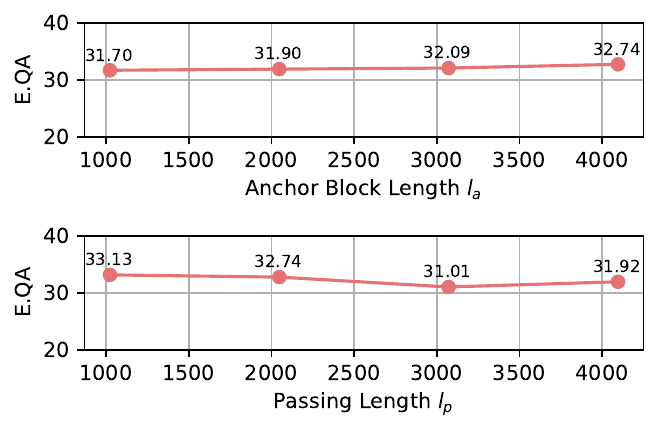}
\end{center}
\caption{The E.QA performance of \name~with various anchor block length $l_a$ and passing length $l_p$. We select the hyperparameters from $\{1024, 2048, 3072, 4096\}$.}
\label{fig:hyperparam}
\end{figure}

\begin{table*}[t]
\small

\centering
\scalebox{0.9}{
\begin{tabular}{l|cccccccccc|c}
\toprule
Method & R.PassKey & R.Number & R.KV &  E.Sum & E.QA & E.MC & E.Dia & Z.QA & C.Debug & M.Find & Avg.\\
\midrule
\multicolumn{12}{c}{\texttt{Llama-3.1-8B-instruct}} \\ \midrule
\textsc{FlashAttn} & 
4056 & 3998 & 3530 & 609 & 3985 & 3924 & 4280 & 4137 & 3959 & 5209 & 3769
\\ 
\textsc{Ulysses} &
30999 & 30426 & 25266 & 3423 & 27616 & 25860 & 29654 & 25332 & 31289 & 39967 & 26983
\\
\textsc{RingAttn}&
20440 & 20218 & 17507 & 3082 & 19476 & 18577 & 20826 & 18779 & 20092 & 25956 & 18495
\\
\textsc{MInference}&
7910 & 7686 & 5259 & 832 & 6186 & 7238 & 7016 & 6399 & 7719 & 8027 & 6427
\\
\textsc{StarAttn} &
33968 & 32632 & 25840 & 2605 & 27896 & 31986 & 32267 & 27266 & 31942 & 37051 & 28345
\\
\rowcolor{pink!20}
\textbf{\name} &
\textbf{44995} & \textbf{43111} & \textbf{32086} & \textbf{2848} & \textbf{32930} & \textbf{41040} & \textbf{36553} & \textbf{32057} & \textbf{44780} & \textbf{45117} & \textbf{35552}
\\
\midrule
\multicolumn{12}{c}{\texttt{Qwen-2.5-14B-instruct}} \\
\midrule
\textsc{FlashAttn} & 
2155 & 2082 & 1736 & 706 & 1929 & 1758 & 1863 & 1882 & 2131 & 2347 & 1859
\\ 
\textsc{Ulysses} &
17214 & 16461 & 12623 & 1574 & 13374 & 13398 & 14092 & 11131 & 16932 & 18460 & 13526
\\
\textsc{RingAttn}&
11056 & 10589 & 8658 & 1314 & 10140 & 9539 & 9982 & 8977 & 10919 & 11676 & 9285
\\
\textsc{MInference}&
3527 & 3305 & 2244 & 618 & 2645 & 2498 & 2480 & 2222 & 3669 & 3785 & 2699
\\
\textsc{StarAttn} &
18793 & 17892 & 13695 & \textbf{1641} & 14604 & 13874 & 14462 & 12453 & 18640 & 19640 & 14569
\\
\rowcolor{pink!20}
\textbf{\name} &
\textbf{25218} & \textbf{23643} & \textbf{17192} & 1592 & \textbf{16875} & \textbf{17937} & \textbf{17267} & \textbf{13303} & \textbf{25134} & \textbf{25929} & \textbf{18409}
\\
\midrule
\multicolumn{12}{c}{\texttt{Yi-34B-200K}} \\
\midrule
\textsc{FlashAttn} & 
1041& 1089& 907& 372& 1110& 1137& 1099& 1125& 1111& 1208& 1030\\ 
\textsc{Ulysses} &
9192& 8820& 7383& 1369& 8523& 8388& 8841& 8374& 9016& 9901& 7981\\
\textsc{RingAttn}&
6203& 6032& 5296& 1281& 6021& 5785& 5951& 6024& 5989& 6552& 5513\\
\textsc{MInference}&
1732& 1637& 1199& 400& 1327& 1760& 1523& 1062& 1766& 1806& 1421\\
\textsc{StarAttn} &
9039& 8675& 7284& 1893& 8419& 9007& 8866& 7877& 9046& 9373& 7948\\
\rowcolor{pink!20}
\textbf{\name} &
\textbf{12441}& \textbf{9404}& \textbf{2059}& \textbf{10798}& \textbf{12726}& \textbf{12233}& \textbf{9327}& \textbf{12744}& \textbf{9328}& \textbf{12744}& \textbf{10624}\\

\bottomrule
\end{tabular}
}
\caption{The inference speed of \name~compared with all the baselines on $\infty$Bench. The ``Avg.'' represents the average speed. The highest score in each column is marked in \textbf{bold}. We report the speed in ``tok/s''.}
\label{tab:InfiniteBench-speed}
\vspace{-10pt}
\end{table*}

The experimental results shown in Figure~\ref{fig:hyperparam} indicate that both $l_a$ and $l_p$ are stable. While there is a slight performance improvement with larger $l_a$, the variation remains insignificant. The performance change with varying $l_p$ is also minimal. This suggests that it is not necessary to tune $l_a$ and $l_p$ delicately, making \name~easy to use.

\subsection{Short-Context Performance of \name}
\label{sec:short-context}

We report the details and hyperparameters of short-context performance experiment here.

As reperted in Section \ref{sec:short-context-main}, we compare task performance and inference speed of RULER using \textsc{FullAttn} (implemented with \textsc{FlashAttn}) and \name, with the context length set to 4K tokens, utilizing \texttt{Llama-3-8B-instruct-1M} as the backbone model. In addition, the number of hosts $H$ is set to 4, and both anchor length $l_a$ and passing length $l_p$ are set to 256. We evaluate each method on 50 samples per task. We align the settings for inference speed measurement with those used in Figure \ref{fig:acc-speed}. The detailed results are reported in Table \ref{tab:shortlen-acc}.

\subsection{Accuracy Experiment on LoCoCo}

To present a comparison of \name~against optimization strategies that target KV cache size reduction to enable faster inference, we select LoCoCo as a baseline here for comparison with \name. We report its performance in this section. 
As Table \ref{tab:lococo} shows, LoCoCo achieves low performance on RULER, as its fixed KV cache size leads to significant information loss, hindering its ability to handle complex long-context tasks effectively. The performance of \textsc{FullAttn} and \name~corresponds to the results reported in Table \ref{tab:results-ruler} of the main text, which were evaluated using the \texttt{Llama-3.1-8B-instruct model}. Meanwhile, for the evaluation of LoCoCo, we test only 20 samples for each task.

\subsection{The Detailed Experimantal Results}
\label{sec:detail-results}
Here, we present the detailed experimental results to extensively illustrate each figure in the main text, Section~\ref{sec:break_pd} and ~\ref{sec:hyper_stability}.
In Section~\ref{sec:end-to-end}, the detailed experimental results of Figure~\ref{fig:acc-speed} are provided in Table~\ref{tab:InfiniteBench-speed} and Table~\ref{tab:RULER-speed}. Table~\ref{tab:InfiniteBench-speed} provides the accurate speed for all methods on $\infty$Bench, while the speed is the average value across all tasks. By contrast, Table~\ref{tab:RULER-speed} reports the precise speed measurements for all methods on RULER, with the values similarly averaged over all tasks.
The performance of each method in Figure~\ref{fig:acc-speed} is provided in Table~\ref{tab:results-infinitebench} and Table~\ref{tab:results-ruler}.
In Section~\ref{sec:varlen}, Figure~\ref{tab:varlen-acc} and Figure~\ref{tab:varlen-speed} present the detailed results corresponding to Figure~\ref{fig:var-seqlen}. Specifically, Table~\ref{tab:varlen-acc} aligns with Figure~\mbox{\ref{fig:var-seqlen}\hspace{-20pt}\subref{fig:var-seqlen-a}} while Table~\ref{tab:varlen-speed} corresponds to Figure~\mbox{\ref{fig:var-seqlen}\hspace{-20pt}\subref{fig:var-seqlen-b}}. 
In Section~\ref{sec:abl}, the detailed experimental results of Figure~\ref{fig:breakdown-tb} are provided in Table~\ref{tab:breakdown-tb}.
In Section~\ref{sec:break_pd}, we add the breakdown analysis of the prefill and decoding wall-time for various methods, while the results are illustrated in Figure~\ref{fig:breakdown-pd}. The detailed experimental results are provided in Table~\ref{tab:breakdown-pd}.
In Section~\ref{sec:intro}, the detailed experimental results of Figure~\ref{fig:pt} are provided in Table~\ref{tab:pt}. The anchor length and passing length are set to 1024 and 512 for 32K input length, 2048 and 1024 for 64K input length, 4096 and 2048 for 128K-1024K input length.

\begin{table}[H]
\small
\center
\scalebox{0.9}{
\begin{tabular}{l|cc}
\toprule
& Prefill Time & Decoding Time \\
\midrule
\textsc{FlashAttn}   &   30137.03&   422.31\\
\textsc{Ulysses}   &   4028.66&   263.52\\
\textsc{RingAttn}   &   6317.03&   185.33\\
\textsc{MInference}   &   18067.32&   2275.61\\
\textsc{StarAttn}   &   3556.60&   208.23\\
\rowcolor{pink!20}
\textbf{\name~}   &   2554.77&   284.21\\
\bottomrule
\end{tabular}}
\caption{The accurate time of Figure~\ref{fig:breakdown-pd}. All the time is reported in ``ms''. We breakdown the wall-time of Transformer inference into prefill and decoding time.}
\label{tab:breakdown-pd}
\end{table}

\begin{table}[H]
\small
\center
\scalebox{0.8}{
\begin{tabular}{l|cccccc}
\toprule
    & 32K & 64K & 128K & 256K & 512K & 1024K \\
\midrule
\textsc{FlashAttn}   & 3.46& 9.51& 30.01& OOM & OOM & OOM \\
\textsc{Ulysses}  & 0.50& 1.30& 3.95& 13.49& 49.55& OOM \\
\textsc{RingAttn}   & 0.72& 2.00& 6.34& 21.80& 81.62& OOM \\
\textsc{MInference}   & 4.95& 8.72& 15.16& OOM & OOM & OOM \\
\textsc{StarAttn}   & 0.67& 1.43& 3.50& 9.63& 30.43& OOM \\
\rowcolor{pink!20}
\textbf{\name~}   & \textbf{0.49} & \textbf{1.09} &  \textbf{2.79} & \textbf{5.53} & \textbf{13.39} &  \textbf{37.60}\\
\bottomrule
\end{tabular}}
\caption{The prefill time of Figure~\ref{fig:pt}. We report the time in ``s''. ``OOM'' represents out-of-memory error.}

\label{tab:pt}
\end{table}

\begin{table*}[t!]
\small

\centering
\scalebox{0.82}{
\begin{tabular}{l|ccccccccccccc|c}
\toprule
Method & SG1 & SG2 & SG3 & MK1 & MK2 & MK3 & MV & MQ & VT & CWE & FWE & QA1 & QA2 & Avg.\\
\midrule
\multicolumn{15}{c}{\texttt{Llama-3.1-8B-instruct}} \\ \midrule
\textsc{FlashAttn}& 
 4277&  4282&  4062&  4317&  4293&  4145&  3908&  3899&  3847&  4203&  4056&  4361&  4382& 
4156\\
\textsc{Ulysses}&
 30138&  29537&  25214&  29611&  29542&  24749&  20721&  20212&  27411&  19994&  21721&  29547&  31609& 
26154\\
\textsc{RingAttn}&
 19642&  19521&  17318&  19760&  19622&  17298&  14582&  15072&  18716&  14731&  16244&  20022&  19862& 
17876\\
\textsc{MInference} &
 7660&  7579&  7555&  7362&  5780&  5201&  4509&  3245&  7720&  7148&  6544&  5604&  3617& 
6117\\
\textsc{StarAttn} &
 31636&  32532&  25900&  32436&  32314&  25378&  17494&  20770&  29319&  22738&  22901&  19696&  33661& 
26675\\
\rowcolor{pink!20}
\textbf{\name} &
\textbf{43930}& \textbf{44504}&\textbf{33106}&\textbf{44502}&\textbf{43717}&\textbf{33882}&\textbf{31328}&\textbf{25349}&\textbf{39680}&\textbf{25580}&\textbf{27828}&\textbf{42931}&\textbf{45664}&\textbf{37077}\\
\midrule
\multicolumn{15}{c}{\texttt{Qwen-2.5-14B-instruct}} \\
\midrule
\textsc{FlashAttn}& 
 2175&  1958&  1819&  2054&  1925&  1438&  1426&  1315&  1882&  1352&  1982&  2038&  2045& 
1801\\
\textsc{Ulysses}&
 16221&  13968&  13761&  15047&  14713&  11345&  10552&  9863&  14312&  11420&  15174&  15642&  15464& 
13652\\
\textsc{RingAttn}&
 10571&  9558&  9322&  10070&  9892&  7771&  7791&  7487&  9816&  8267&  10505&  10464&  10358& 
9375\\
\textsc{MInference} &
 3391&  2097&  2637&  2408&  2389&  1476&  1845&  1634&  2721&  1465&  2756&  3134&  3010& 
2382\\
\textsc{StarAttn} &
 18226&  16314&  15403&  18285&  15941&  11292&  13469&  9292&  15276&  11039&  16059&  16118&  15965& 
14821\\
\rowcolor{pink!20}
\textbf{\name} &
\textbf{24576}& \textbf{21588}&\textbf{19580}&\textbf{20732}&\textbf{23305}&\textbf{14292}&\textbf{15638}&\textbf{10786}&\textbf{18601}&\textbf{13469}&\textbf{19301}&\textbf{21758}&\textbf{20339}&\textbf{18767}\\
\midrule
\multicolumn{15}{c}{\texttt{Yi-34B-200K}} \\
\midrule
\textsc{FlashAttn}& 
 1194&  1146&  1122&  1147&  1182&  1144&  1078&  1099&  1174&  1140&  1224&  1167&  1186& 
1154\\
\textsc{Ulysses}&
 8466&  5513&  4427&  5110&  8659&  7498&  6144&  6674&  8152&  7083&  8099&  8130&  8130& 
7083\\
\textsc{RingAttn}&
 4478&  3895&  3915&  5158&  5814&  5286&  4657&  4862&  5615&  5092&  5685&  5620&  5625& 
5054\\
\textsc{MInference} &
 1644&  1279&  1199&  1340&  1585&  1293&  1012&  1101&  1450&  1210&  1361&  1484&  1461& 
1340\\
\textsc{StarAttn} &
 8554&  7569&  6939&  7690&  8713&  7442&  6933&  6704&  8108&  7086&  7826&  8160&  8151& 
7683\\
\rowcolor{pink!20}
\textbf{\name} &
\textbf{11989}& \textbf{10176}&\textbf{8902}&\textbf{10030}&\textbf{12061}&\textbf{9768}&\textbf{7910}&\textbf{8396}&\textbf{10252}&\textbf{9135}&\textbf{10264}&\textbf{10990}&\textbf{10777}&\textbf{10050}\\

\bottomrule
\end{tabular}
}
\caption{The inference speed of \name~compared with all the baselines on RULER. The ``Avg.'' represents the average speed. The highest score in each column is marked in \textbf{bold}. We report the speed in ``tok/s''.}
\label{tab:RULER-speed}
\end{table*}

\begin{table*}[h]
\small
\center
\scalebox{0.85}{
\begin{tabular}{l|ccccccc|c}
\toprule
    & QKV Projection & Retaining Head & Communication & Attention & O Projection & FFN & Others & Transformer Block \\
\midrule
\textsc{FlashAttn}   & 25.33& -- & -- & 664.01& 17.42& 201.44& 32.67& 940.86\\
\textsc{Ulysses}   & 3.31& -- & 3.90& 84.53& 2.27& 25.88& 4.62& 124.51\\
\textsc{RingAttn}   & 3.21& -- & 18.40& 152.12& 2.09& 24.40& 4.62& 205.19\\
\textsc{MInference}   & 24.45& -- & -- & 281.39& 15.88& 201.56& 40.80& 564.07\\
\textsc{StarAttn}   & 6.67& -- & -- & 41.84& 4.29& 50.01& 8.56& 111.37\\
\rowcolor{pink!20}
\textbf{\name}   & 4.01& 1.72& 0.62& 34.07& 2.67& 30.76& 6.33& 80.18\\
\bottomrule
\end{tabular}}
\caption{The accurate time of Figure~\ref{fig:breakdown-tb}. All the time is reported in ``ms''. We breakdown the wall-time of each Transformer block into 7 components. ``--'' indicates that the time of this component does not exist in the corresponding method.}

\label{tab:breakdown-tb}
\end{table*}

\begin{table*}[t!]
\small

\centering
\scalebox{0.82}{
\begin{tabular}{l|ccccccccccccc|c}
\toprule
Method & SG1 & SG2 & SG3 & MK1 & MK2 & MK3 & MV & MQ & VT & CWE & FWE & QA1 & QA2 & Avg.\\
\midrule
\multicolumn{15}{c}{$n=32$K} \\ \midrule
\textsc{FullAttn} & 
100.00& 100.00& 98.00& 100.00& 96.00& 82.00& 97.00& 98.50& 92.00& 40.20& 88.00& 82.00& 64.00& 87.52\\ \midrule
\textsc{MInference} &
100.00& 100.00& 100.00& 100.00& 96.00& 76.00& 95.50& 99.00& 90.40& 59.40& 88.00& 80.00& 62.00& 88.18\\
\textsc{StarAttn} &
100.00& 100.00& 100.00& 96.00& 98.00& 96.00& 83.50& 93.50& 88.80& 76.20& 90.67& 78.00& 62.00& 89.44\\
\rowcolor{pink!20}
\textbf{\name} &
100.00& 98.00& 100.00& 100.00& 98.00& 100.00& 89.50& 98.50& 86.40& 78.00& 90.00& 74.00& 60.00& 90.18\\
\midrule
\multicolumn{15}{c}{$n=64$K} \\
\midrule
\textsc{FullAttn} & 
100.00& 100.00& 98.00& 100.00& 98.00& 56.00& 99.00& 98.00& 84.40& 1.20& 78.67& 68.00& 54.00& 79.64\\ \midrule
\textsc{MInference} &
100.00& 100.00& 100.00& 100.00& 98.00& 54.00& 97.50& 99.50& 78.00& 6.20& 81.33& 64.00& 58.00& 79.73\\
\textsc{StarAttn} &
100.00& 94.00& 100.00& 96.00& 96.00& 86.00& 81.50& 94.50& 82.40& 16.00& 82.00& 70.00& 52.00& 80.80\\
\rowcolor{pink!20}
\textbf{\name} &
100.00& 96.00& 100.00& 98.00& 98.00& 100.00& 95.00& 97.50& 80.40& 11.40& 85.33& 72.00& 58.00& 83.97\\
\midrule
\multicolumn{15}{c}{$n=128$K} \\
\midrule
\textsc{FullAttn} & 
100.00 & 100.00 & 100.00& 98.00& 100.00& 36.00& 98.50& 95.50& 77.20& 0.00& 72.00& 68.00& 46.00& 76.25\\ \midrule
\textsc{MInference} &
100.00& 100.00& 100.00& 100.00& 100.00& 46.00& 96.50& 99.00& 74.80& 0.20& 80.00& 76.00& 54.00& 78.96\\
\textsc{StarAttn} &
100.00& 100.00& 100.00& 96.00& 96.00& 90.00& 79.50& 90.50& 84.40& 0.20& 72.00& 68.00& 52.00& 79.12\\
\rowcolor{pink!20}
\textbf{\name} &
100.00& 100.00& 100.00& 98.00& 94.00& 98.00& 97.00& 98.00& 73.20& 0.20& 70.67& 68.00& 52.00& 80.70\\
\midrule
\multicolumn{15}{c}{$n=256$K} \\
\midrule
\textsc{FullAttn} & 
100.00 & 100.00 & 96.00& 94.00& 97.22& 22.00& 92.50& 95.00& 64.00& 0.60& 76.67& 78.00& 44.00& 73.85\\
\midrule
\textsc{MInference} &
100.00& 100.00& 100.00& 96.00& 94.00& 46.00& 91.85& 95.50& 77.20& 0.20& 79.33& 72.00& 48.00& 76.93\\
\textsc{StarAttn} &
100.00& 96.00& 100.00& 94.00& 86.00& 66.00& 78.50& 83.00& 70.00& 0.20& 80.67& 68.00& 42.00& 74.18\\
\rowcolor{pink!20}
\textbf{\name} &
100.00 & 96.00& 100.00& 94.00& 94.00& 94.00& 93.00& 98.00& 47.60& 0.40& 78.00& 70.00& 46.00& 77.77\\\midrule
\multicolumn{15}{c}{$n=512$K} \\
\midrule
\textsc{FullAttn} & 
98.00& 98.00& 100.00& 94.00& 76.00 & 10.00& 90.50& 96.00& 46.80& 0.60& 86.67& 70.00& 46.00& 70.20\\
\midrule
\textsc{MInference} &
100.00& 100.00& 100.00& 98.00& 76.00& 12.00& 85.50& 91.00& 77.20& 0.40& 83.33& 62.00& 42.00& 71.34\\
\textsc{StarAttn} &
100.00& 98.00& 100.00& 88.00& 74.00& 20.00& 71.50& 80.00& 45.20& 0.40& 88.00& 60.00& 38.00& 66.39\\
\rowcolor{pink!20}
\textbf{\name} &
100.00& 100.00& 100.00& 90.00& 88.80& 80.00& 88.80& 95.00& 25.20& 0.80& 83.33& 72.00& 44.00& 74.33\\
\bottomrule
\end{tabular}
}
\vspace{-5pt}
\caption{The task performance of \name~compared with all the baselines on RULER across different input length $n$, where higher score represents better performance. ``Avg.'' represents the average score. }
\label{tab:varlen-acc}
\end{table*}

\begin{table*}[t!]
\small

\centering
\scalebox{0.82}{
\begin{tabular}{l|ccccccccccccc|c}
\toprule
Method & SG1 & SG2 & SG3 & MK1 & MK2 & MK3 & MV & MQ & VT & CWE & FWE & QA1 & QA2 & Avg.\\
\midrule
\multicolumn{15}{c}{$n=32$K} \\ \midrule
\textsc{FlashAttn} & 
8955 & 8764 & 8754 & 7369 & 8818 & 7534 & 6259 & 6751 & 8005 & 6614 & 8200 & 8333 & 8121 & 7883\\ 
\textsc{Ulysses} & 
42479 & 40140 & 22599 & 39735 & 45166 & 23239 & 14300 & 15644 & 27447 & 15465 & 26981 & 32685 & 25030 & 28532\\ 
\textsc{RingAttn} & 
31432 & 32855 & 16703 & 29762 & 32016 & 19901 & 14227 & 15243 & 21899 & 12687 & 20532 & 26173 & 22230 & 22743\\ 
\textsc{MInference} &
4940 & 4714 & 3116 & 4644 & 4725 & 3018 & 1771 & 2097 & 3765 & 2094 & 3608 & 3874 & 3657 & 3540\\
\textsc{StarAttn} &
33909 & 34256 & 17630 & 34951 & 36137 & 20604 & 14362 & 12704 & 23505 & 14008 & 23792 & 27637 & 24730 & 24479\\
\rowcolor{pink!20}
\textbf{\name} &
50778 & 43160 & 18264 & 41896 & 45641 & 21149 & 24893 & 13909 & 30407 & 13457 & 25016 & 29156 & 30863 & 29892\\
\midrule
\multicolumn{15}{c}{$n=64$K} \\
\midrule
\textsc{FlashAttn} & 
6647 & 6664 & 6026 & 6606 & 6627 & 6208 & 5006 & 5526 & 6334 & 5619 & 6411 & 6344 & 6468 & 6191\\ 
\textsc{Ulysses} & 
38672 & 37904 & 22892 & 36890 & 38257 & 24925 & 13190 & 18511 & 28011 & 17790 & 27585 & 28760 & 30551 & 27995\\ 
\textsc{RingAttn} & 
27484 & 27209 & 20289 & 27089 & 27621 & 21167 & 11823 & 16726 & 22120 & 15897 & 22340 & 23076 & 23959 & 22062\\ 
\textsc{MInference} &
5531 & 5437 & 3954 & 5527 & 5156 & 4012 & 2069 & 2531 & 4609 & 3130 & 4364 & 4539 & 4868 & 4287\\
\textsc{StarAttn} &
35871 & 35840 & 21658 & 33590 & 33620 & 23213 & 17342 & 16487 & 28760 & 19524 & 27363 & 26143 & 26949 & 26643\\
\rowcolor{pink!20}
\textbf{\name} &
48822 & 46944 & 32005 & 50754 & 51244 & 29646 & 19554 & 19655 & 30925 & 21003 & 28469 & 29516 & 30056 & 33738\\
\midrule
\multicolumn{15}{c}{$n=128$K} \\
\midrule
\textsc{FlashAttn} & 
4262 & 4233 & 3984 & 4285 & 4209 & 4062 & 3463 & 3766 & 4161 & 3910 & 4330 & 4184 & 4266 & 4086\\ 
\textsc{Ulysses} & 
30195 & 29300 & 25446 & 30020 & 30070 & 25720 & 16690 & 20717 & 28030 & 22484 & 28064 & 26920 & 26951 & 26200\\ 
\textsc{RingAttn} & 
20013 & 19221 & 17226 & 19559 & 19565 & 17264 & 13162 & 15344 & 18501 & 16042 & 18622 & 18715 & 18447 & 17822\\ 
\textsc{MInference} &
5389 & 5385 & 4459 & 5427 & 5097 & 4597 & 2275 & 3167 & 5149 & 3762 & 4851 & 4789 & 4734 & 4545\\
\textsc{StarAttn} &
34734 & 34416 & 29086 & 34171 & 33984 & 28468 & 19095 & 20370 & 31682 & 25456 & 31313 & 31339 & 30678 & 29600\\
\rowcolor{pink!20}
\textbf{\name} &
46644 & 46110 & 36421 & 42653 & 46052 & 35107 & 18480 & 27517 & 40831 & 30995 & 39476 & 40254 & 37931 & 37575\\
\midrule
\multicolumn{15}{c}{$n=256$K} \\
\midrule
\textsc{FlashAttn} & 
OOM & OOM & OOM & OOM & OOM & OOM & OOM & OOM & OOM & OOM & OOM & OOM & OOM & OOM\\
\textsc{Ulysses} & 
18365 & 18295 & 16987 & 18385 & 18265 & 17231 & 12866 & 15530 & 17689 & 16117 & 17894 & 17710 & 17735 & 17159\\ 
\textsc{RingAttn} & 
11847 & 11748 & 11335 & 11777 & 11849 & 11431 & 9325 & 10654 & 11642 & 10925 & 11799 & 11661 & 11663 & 11358\\ 
\textsc{MInference} &
5046 & 5223 & 4262 & 5027 & 4744 & 4411 & 2315 & 2711 & 4930 & 3705 & 4682 & 4519 & 4559 & 4318\\
\textsc{StarAttn} &
26455 & 26174 & 23985 & 26096 & 26304 & 24073 & 18374 & 18399 & 25383 & 22303 & 24674 & 25413 & 24953 & 24045\\
\rowcolor{pink!20}
\textbf{\name} &
34126 & 33613 & 29983 & 34000 & 33914 & 30404 & 19710 & 25390 & 32728 & 26596 & 31437 & 31574 & 32469 & 30457\\
\midrule
\multicolumn{15}{c}{$n=512$K} \\
\midrule
\textsc{FlashAttn} & 
OOM & OOM & OOM & OOM & OOM & OOM & OOM & OOM & OOM & OOM & OOM & OOM & OOM & OOM\\ 
\textsc{Ulysses} & 
10327 & 10249 & 10089 & 10299 & 10317 & 9993 & 8910 & 9334 & 10324 & 9797 & 10458 & 10192 & 10259 & 10042\\ 
\textsc{RingAttn} & 
6395 & 6337 & 6246 & 6331 & 6383 & 6253 & 5783 & 5937 & 6342 & 6138 & 6508 & 6332 & 6332 & 6255\\ 
\textsc{MInference} &
4796 & 4720 & 3999 & 4468 & 4465 & 4134 & 2261 & 2314 & 4585 & 3613 & 4562 & 4367 & 4332 & 4047\\
\textsc{StarAttn} &
16841 & 16661 & 16130 & 16556 & 16746 & 15988 & 13936 & 14365 & 16969 & 15737 & 16969 & 16798 & 16600 & 16177\\
\rowcolor{pink!20}
\textbf{\name} &
28100 & 27614 & 25938 & 28315 & 28306 & 25773 & 19435 & 22188 & 27541 & 24470 & 27484 & 27087 & 27082 & 26102\\
\bottomrule
\end{tabular}
}
\caption{The inference speed of \name~compared with all the baselines on RULER across different input length $n$. ``Avg.'' represents the average speed. We report the speed in ``tok/s''. ``OOM'' represents out-of-memory error.}
\label{tab:varlen-speed}
\end{table*}

\begin{table*}[t!]
\small

\centering
\scalebox{0.78}{
\begin{tabular}{l|ccccccccccccc|c|c}
\toprule
Method & SG1 & SG2 & SG3 & MK1 & MK2 & MK3 & MV & MQ & VT & CWE & FWE & QA1 & QA2 & Avg.& tok/s.\\
\midrule
\textsc{FullAttn} & 
\textbf{100.00}& \textbf{98.00}& \textbf{100.00}& \textbf{98.00}& 98.00& \textbf{100.00}& \textbf{100.00}& 97.00& \textbf{98.80}& 97.20& 94.00& 84.00& 64.00& 94.54& 6247.58\\ 
\rowcolor{pink!20}
\textsc{\name~} &
\textbf{100.00}& 92.00& \textbf{100.00}& \textbf{98.00}& \textbf{100.00}& 94.00& \textbf{100.00}& \textbf{99.50}& \textbf{98.80}& \textbf{98.00}& \textbf{95.33}& \textbf{90.00}& \textbf{68.00}& \textbf{94.89}& \textbf{6597.47} \\
\bottomrule
\end{tabular}
}
\caption{The task performance and inference speed of \name~and \textsc{FullAttn} (\textsc{FlashAttn}) on RULER, where input length is set to 4K tokens. We report the speed in  ``tok/s''. ``Avg.'' represents the average score.}
\label{tab:shortlen-acc}
\end{table*}

\begin{table*}[t!]
\small

\centering
\scalebox{0.82}{
\begin{tabular}{l|ccccccccccccc|c}
\toprule
Method & SG1 & SG2 & SG3 & MK1 & MK2 & MK3 & MV & MQ & VT & CWE & FWE & QA1 & QA2 & Avg\\
\midrule
\textsc{FullAttn} & 
99.40& 99.80& 99.60& \textbf{98.20}& 87.60& 67.00& 94.65& \textbf{98.00}& \textbf{60.98}& \textbf{71.40}& 72.20& \textbf{78.20}& \textbf{41.60}& \textbf{82.20}\\
\textsc{LoCoCo} & 
30.00& 10.00& 0.00& 0.00& 10.00& 10.00& 5.00& 7.50& 24.00& 56.50& 75.00& 40.00& 20.00& 22.15\\
\rowcolor{pink!20}
\textsc{\name~} &
\textbf{100.00}& \textbf{100.00}& \textbf{99.80}& 85.60& \textbf{91.00}& \textbf{89.00}& \textbf{95.05}& 96.40& 51.96&  63.82& \textbf{77.33}& 70.00& 41.20& 81.63 \\
\bottomrule
\end{tabular}
}
\caption{The task performance of LoCoCo compared with \name~and FullAttn on RULER. ``Avg.'' represents the average score. \textsc{FullAttn}
represents \textsc{FlashAttn}, \textsc{RingAttn}, and \textsc{Ulysses}, as their computational results remain unchanged.}
\label{tab:lococo}
\end{table*}

\end{document}